\documentclass[10pt,journal,compsoc]{IEEEtran}

\usepackage{times}
\usepackage{epsfig}
\usepackage{graphicx}
\usepackage{amsmath}
\usepackage{amssymb}
\usepackage{bm}
\usepackage{overpic}
\usepackage{color}
\usepackage{multirow}
\usepackage{booktabs}       % professional-quality tables
\usepackage{array}
\usepackage{paralist}
\usepackage[colorlinks,
            linkcolor=red,
            anchorcolor=blue,
            citecolor=green
            ]{hyperref}
\usepackage{url}
\newcommand{\PreserveBackslash}[1]{\let\temp=\\#1\let\\=\temp}
\newcolumntype{C}[1]{>{\PreserveBackslash\centering}p{#1}}
\newcolumntype{R}[1]{>{\PreserveBackslash\raggedleft}p{#1}}
\newcolumntype{L}[1]{>{\PreserveBackslash\raggedright}p{#1}}
\newcommand{\Paragraph}[1]{\noindent\textbf{#1}}
\newcommand{\mlr}[1]{\multirow{2}{*}{#1}}

\ifCLASSOPTIONcompsoc
  \usepackage[nocompress]{cite}
\else
  \usepackage{cite}
\fi

\def\etal{\emph{et al}.}
\def\ie{i.e.,~} % that is, in other words
\def\eg{\emph{e.g}}

\def\red#1{\textbf{\color{red} #1}} %in Table
\def\blu#1{\textbf{\color{blue}\underline{#1}}} %in Table

\ifCLASSINFOpdf
\else
\fi

\begin{document}
\title{PiCANet: Pixel-wise Contextual Attention Learning for Accurate Saliency Detection}

\author{Nian~Liu,
        Junwei~Han,~\IEEEmembership{Senior~Member,~IEEE,}
        and~Ming-Hsuan~Yang,~\IEEEmembership{Fellow,~IEEE}% <-this % stops a space
\IEEEcompsocitemizethanks{\IEEEcompsocthanksitem N. Liu and J. Han are with School of Automation, Northwestern Polytechnical University, China. E-mail: \{liunian228,junweihan2010\}@gmail.com
\IEEEcompsocthanksitem M.-H. Yang is with School of Engineering, University of California, Merced, CA, US. E-mail: mhyang@ucmerced.edu
}
% \thanks{Manuscript received April 19, 2005; revised August 26, 2015.}
}

\markboth{Journal of \LaTeX\ Class Files,~Vol.~14, No.~8, August~2015}%
{Shell \MakeLowercase{\textit{et al.}}: Bare Demo of IEEEtran.cls for Computer Society Journals}

\IEEEtitleabstractindextext{%
\begin{abstract}
 In saliency detection, every pixel needs contextual information to make saliency prediction. Previous models usually incorporate contexts holistically. However, for each pixel, usually only part of its context region is useful and contributes to its prediction, while some other part may serve as noises and distractions. In this paper, we propose a novel pixel-wise contextual attention network, \ie PiCANet, to learn to selectively attend to informative context locations at each pixel. Specifically, PiCANet generates an attention map over the context region of each pixel, where each attention weight corresponds to the relevance of a context location w.r.t the referred pixel. Then, attentive contextual features can be constructed via selectively incorporating the features of useful context locations with the learned attention. We propose three specific formulations of the PiCANet via embedding the pixel-wise contextual attention mechanism into the pooling and convolution operations with attending to global or local contexts. All the three models are fully differentiable and can be integrated with CNNs with joint training. We introduce the proposed PiCANets into a U-Net \cite{ronneberger2015unet} architecture for salient object detection. Experimental results indicate that the proposed PiCANets can significantly improve the saliency detection performance. The generated global and local attention can learn to incorporate global contrast and smoothness, respectively, which help localize salient objects more accurately and highlight them more uniformly. Consequently, our saliency model performs favorably against other state-of-the-art methods. Moreover, we also validate that PiCANets can also improve semantic segmentation and object detection performances, which further demonstrates their effectiveness and generalization ability.
\end{abstract}

}

\maketitle

\IEEEraisesectionheading{\section{Introduction}\label{sec:introduction}}

\IEEEPARstart{S}{aliency} detection mimics the human visual attention mechanism to highlight distinct regions or objects that catch people's eyes in visual scenes. It is one of the most basic low-level image preprocessing techniques and can benefit lots of high-level image understanding tasks, such as image classification \cite{sharma2012discriminative}, object detection \cite{zhang2018leveraging}, and semantic segmentation \cite{wei2017stc,chaudhry_dcsp_2017}.
%
%In computer vision community, saliency detection involves two branches: eye fixation prediction which focuses on estimating probable human eye attended locations in images, and salient object detection which aims at accurately and uniformly highlighting the most salient objects in visual scenes.
%%
%This paper mainly focuses on the second branch.

Contextual information plays a crucial role in saliency detection, which is typically reflected in the contrast mechanism. In one of the earliest pioneering work \cite{itti1998model}, Itti \emph{et al}. propose to compute the feature difference between each pixel and its surrounding regions in a Gaussian pyramid as contrast. Following this idea, many subsequent models \cite{han2011bottom,cheng2015global,klein2011center} also employ the contrast mechanism to model visual saliency. In these methods, local context or global context is utilized as the reference to evaluate the contrast of each image pixel, which is referred as the local contrast or the global contrast, respectively. Generally, a feature representation is first extracted for each image pixel. Then the features of all the referred contextual locations are aggregated into an overall representation as the contextual feature to infer contrast.
% via various aggregation schemes, such as feature histogram in \cite{cheng2015global,klein2011center}, sparse coding in \cite{han2013object}, and average pooling in \cite{jiang2013drfi,wang2015legs}.

%Recently, convolutional neural networks (CNNs) have achieved impressive successes in various computer vision tasks (\eg, image classification \cite{krizhevsky2012alexnet,simonyan2014vgg,he2016resnet}, object detection \cite{girshick2014rcnn,liu2016ssd}, and semantic segmentation \cite{long2015fcn,chen2018deeplab,ronneberger2015unet}) due to their extraordinary capability to learn discriminative feature representations with deep architectures. Following these successful cases, several work also adopt CNNs in saliency detection.
Recently, following the impressive successes that convolutional neural networks (CNNs) have achieved on other vision tasks (\eg, image classification \cite{simonyan2014vgg}, object detection \cite{girshick2014rcnn,liu2016ssd}, and semantic segmentation \cite{long2015fcn,chen2018deeplab,ronneberger2015unet}), several work also adopt CNNs in saliency detection. Specifically, early work \cite{li2015mdf,liu2018mrcnn,zhao2015mcdl} usually use CNNs in a sliding window fashion, where for each pixel or superpixel deep features are first extracted from multiscale image regions and then are combined to infer saliency. Recent work \cite{kuen2016recurrent,li2016dcl,liu2016dhsnet,wang2016rfcn,hou2017dss,luo2017nldf,Zhang2017amulet,Wang2017srm,liu2018dsclrcn,wang2018asnet,zhang2018bmp,wang2018dgrl,zhang2018pagrn,li2018c2snet,chen2018ra} typically extract multilevel convolutional features from fully convolutional networks (FCNs) \cite{long2015fcn} and subsequently use various neural network modules to infer saliency by combining the feature maps. As for contextual features, the first school models compute them from corresponding input image regions, while the second school methods obtain them from corresponding receptive fields of different convolutional layers. Most of them utilize the entire context regions to construct the contextual features due to the fixed intrinsic structure of CNNs.

\begin{figure}[!t]
  %\graphicspath{{Figures/figure1/}}
  \centering
  \includegraphics[width=0.8\linewidth]{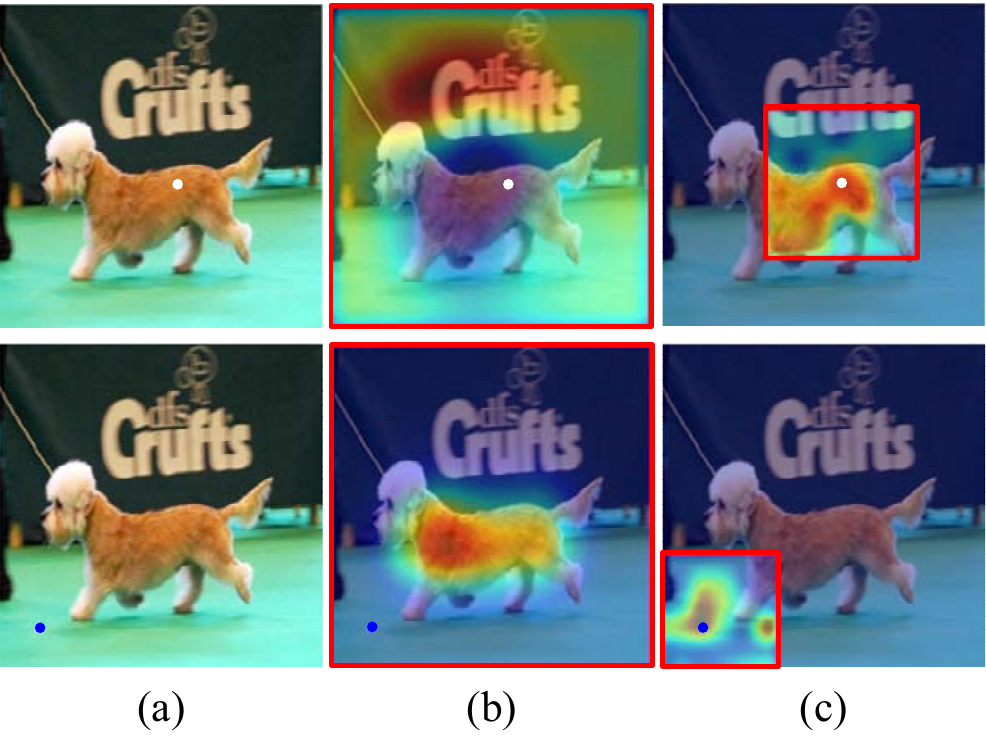}
  \caption{\textbf{Generated global and local pixel-wise contextual attention maps for two example pixels.} (a) shows the original image with two example pixels, in which the white one locates on the foreground dog while the blue one locates on the background. (b) and (c) show their generated global and local contextual attention maps, respectively. Hot color indicates large attention weights. For each case, the referred context region is given by the red box.}
  \label{figure1}
  \vspace{-0.2cm}
\end{figure}

Although in most previous models, including both traditional and CNN based ones, context regions are holistically leveraged, \ie every context location contributes to the contextual feature construction, intuitively this is a sub-optimal choice since every pixel has both useful and useless context parts. In a given context region of a specific pixel, some of its context locations contain relevant information and contribute to its final prediction, while some others are irrelevant and may serve as distractions. Here we give an intuitive example in Figure~\ref{figure1}. For the white dot on the foreground dog in the top row, we can infer its global contrast by comparing it with the background regions. While by referring to the other parts of the dog, we can also conclude that this pixel is part of the foreground dog thus we can uniformly recognize the whole body of the dog. Similarly, for the blue dot in the second row, we can infer its global contrast and affiliation via considering the foreground dog and other regions of the background, respectively. From this example, we can see that when humans are making the prediction for each pixel, we usually refer to its relevant context regions, instead of considering all of them holistically. By doing this, we can learn informative contextual knowledge and thus make more accurate prediction. However, since limited by their model architectures, most existing methods can not address this problem.

To this end, we propose a novel Pixel-wise Contextual Attention Network (PiCANet) to automatically learn to select useful context regions for each image pixel, which is a significant extension from the traditional soft attention model \cite{bahdanau2015attention} to the novel pixel-wise attention concept. Specifically, the proposed PiCANet simultaneously generates every pixel an attention map over its context region. In such way, the relevance of the context locations w.r.t. the referred pixel are encoded in corresponding attention weights. Then for each pixel, we can use the weights to selectively aggregate the features of its context locations and obtain an attentive contextual feature. Thus our model only incorporates useful contextual knowledge and depress other noisy and distractive information for each pixel, which will significantly benefit their prediction. The examples in Figure~\ref{figure1} illustrate different attention maps generated by our model for different pixels.

We design three forms of the PiCANet with contexts of different scopes and the usage of different attention mechanisms. The first two are that we use weighted average to pool global and local contextual features, respectively, linearly aggregating the contexts. We refer them as \emph{global attention pooling} (GAP) and \emph{local attention pooling} (LAP), respectively. The third one is that we adopt the local attention weights in the convolution operation to control the information flow for the convolutional feature extraction at each pixel. We refer this form of the PiCANet as \emph{attention convolution} (AC). All of the three kinds of PiCANet are fully differentiable and can be flexibly embedded into CNNs with end-to-end training.

Based on the proposed PiCANets, we construct a novel network by hierarchically embedding them into a U-Net \cite{ronneberger2015unet} architecture for salient object detection. To be specific, we progressively adopt global and local PiCANets in different decoder modules with multiscale feature maps, thus constructing attentive contextual features with varying context scopes and scales. As a result, saliency inference can be facilitated from these enhanced features. We show some generated attention maps as examples in Figure~\ref{figure1}. The global attention in (b) indicates that it generally follows the global contrast mechanism, \ie the attention of foreground pixels (\eg, the white dot) mainly activate on background regions and vice verse for background pixels (\eg, the blue dot). Figure~\ref{figure1}(c) shows that the local attention generally attends to the visually coherent regions of the center pixel, which can make the predicted saliency maps more smooth and uniform. Besides saliency detection, we also validate the effectiveness of the proposed PiCANets on semantic segmentation and object detection based on widely used baseline models. The results demonstrate that PiCANets can be used as general neural network modules to benefit other dense prediction vision tasks.

In conclusion, we summarize our contributions as follows:

1. We propose novel PiCANets to select informative context regions for each pixel. By using the generated attention weights, attentive contextual features can be constructed for each pixel to facilitate its final prediction.

2. We design three formulations for PiCANets with different context scopes and attention mechanisms. The pixel-wise contextual attention mechanism is introduced into pooling and convolution operations with attention over global or local contexts. Furthermore, all these three formulations are fully differentiable and can be embedded into CNNs with end-to-end training.

3. We embed the PiCANets into a U-Net architecture to hierarchically incorporate attentive global and multiscale local contexts for salient object detection. Experimental results demonstrate that PiCANets can effectively improve the saliency detection performance. As a result, our saliency model performs favorably against other state-of-the-art methods. We also visualize the generated attention maps and find that the global attention generally learns global contrast while the local attention mainly consolidates the smoothness of the saliency maps.

4. We also evaluate the proposed PiCANets on semantic segmentation and object detection. The experimental results further demonstrate the effectiveness and generalization ability of the proposed PiCANets with application on other dense prediction vision tasks.

A preliminary version of this work was published on \cite{liu2018picanet}. In this paper, we mainly make the following extensions with significant improvements. First, based on the two forms of the PiCANet proposed in \cite{liu2018picanet}, we further propose the third formulation by introducing the pixel-wise contextual attention into the convolution operation to modulate the information flow. Experimental results show that it can bring more performance gains for saliency detection. Second, we propose to add explicit supervision for the learning of global attention, which can help to learn global contrast better and improve the model performance. Third, our new model obtains better results than \cite{liu2018picanet} and performs favorably against other state-of-the-art methods published very recently. Forth, we also conduct evaluation experiments on semantic segmentation and object detection to further validate the effectiveness and generalization ability of the proposed PiCANets.

\section{Related Work}

\subsection{Attention Networks}

Attention mechanism is recently introduced into neural networks to learn to select useful information and depress other noises and distractions. Mnih \etal \cite{mnih2014attention} propose a hard attention model trained with reinforcement learning to select discriminative local regions for image classification. Bahdanau \etal \cite{bahdanau2015attention} propose a soft attention model to softly search keywords from the source sentence in machine translation, where the attention model is differentiable thus can be easily trained. Following these work, recently attention models are also applied to various computer vision tasks. In \cite{xu2015show}, Xu \emph{et al}. adopt a recurrent attention model to find relevant image regions for image captioning. Sermanet \etal \cite{sermanet2014attention} propose to select discriminative image regions for fine-grained classification via a recurrent attention model. Similarly, some visual question answering models \cite{xu2016ask,yang2016stacked} also use attention networks to extract features from question-related image regions. In object detection, Li \etal \cite{li2017attentive} adopt an attention model to incorporate target-related regions in global context for optimizing object classification. In \cite{wang2017residual}, spatial attention is learned for each feature map to modulate the features of different spatial locations in the image classification task. In these work, attention models are demonstrated to be helpful in learning more discriminative feature representations via finding informative image regions, thus can benefit the final prediction. Nevertheless, these models only generate one attention map over the whole image (or generate one attention map at each time in a recurrent model), which means they are only optimized to make a single prediction. We refer these attention models as \emph{image-wise contextual attention}.

For dense prediction tasks (\eg, semantic segmentation, and saliency detection), intuitively it is better to generate one attention map for each pixel since making different predictions at different pixels needs different knowledge. Typically, for semantic segmentation, Chen \etal \cite{chen2016attention2scale} first extract multiscale feature maps. Then for each pixel, they generate a set of attention weights to select the optimal scale at that pixel. We refer their attention model as \emph{pixel-wise scale attention}. Whereas we are the first to propose the pixel-wise contextual attention concept for selecting useful context regions for each pixel.

\subsection{Saliency Detection with Deep Learning}

Recently, many saliency detection models adopt various deep networks and achieve promising results. In \cite{liu2015mrcnn,li2015mdf}, sliding windows are used to extract multiple CNN features over cropped multiscale image regions for each pixel or superpixel. Then these features are combined to infer saliency. Zhao \etal \cite{zhao2015mcdl} use a similar idea to combine global and local contextual features. Li \etal \cite{li2016dcl} fuse an FCN based model with a sliding window based model. In \cite{wang2016rfcn}, Wang \emph{et al}. propose to adopt FCNs in a recurrent architecture to progressively refine saliency maps. Liu and Han \cite{liu2016dhsnet} use a U-Net based network to first predict a coarse saliency map from the global view and then hierarchically conduct refinement with finer local features. Similarly, the work in \cite{Zhang2017amulet,luo2017nldf,Wang2017srm,wang2018dgrl,zhang2018pagrn,chen2018ra} also integrate multiscale contextual features for saliency detection with various decoder modules. Hou \etal \cite{hou2017dss} propose a saliency detection model based on the HED network \cite{xie2015hed} where they introduce short connections into the multilevel side outputs. Zhang \etal \cite{zhang2018bmp} design a bi-directional architecture with message passing among different feature maps. In \cite{wang2018asnet}, Wang \emph{et al}. propose to first predict eye fixations and then use them to guide the detection of salient objects. Li \etal \cite{li2018c2snet} use one deep network to perform contour detection and saliency detection simultaneously, and alternately use these two tasks to guide the training of each other.

Most of these models use diverse network architectures to combine multilevel global or local contexts for saliency detection. However, they usually utilize the contexts holistically, without distinguishing useful and other context regions, \eg \cite{liu2016dhsnet,luo2017nldf,Zhang2017amulet}. On the contrary, we propose PiCANets to only incorporate informative context regions. There are also other saliency models adopting attention mechanisms. Specifically, Kuen \etal \cite{kuen2016recurrent} employ a recurrent attention model to select a local region and refine its saliency map at each time step. Zhang \etal \cite{zhang2018pagrn} generate spatial and channel attention for each feature map. In \cite{wang2018dgrl,chen2018ra}, a spatial attention map is adopted for each decoding module to weight the feature map. These models all generate attention once for the whole feature map in each decoding module, thus they still fall into the \emph{image-wise attention} category. While our proposed PiCANets simultaneously generate one contextual attention map for each pixel, thus are more suitable for the dense prediction nature of saliency detection.

\section{Pixel-wise Contextual Attention Network}

\begin{figure*}[!t]
  \graphicspath{{Figures/PiCANet/}}
  \centering
  %\includegraphics[width=1\linewidth]{PiCANet.pdf}
  %\begin{overpic}[width=1\linewidth,grid,tics=1]{PiCANet.pdf}
  \begin{overpic}[width=1\linewidth]{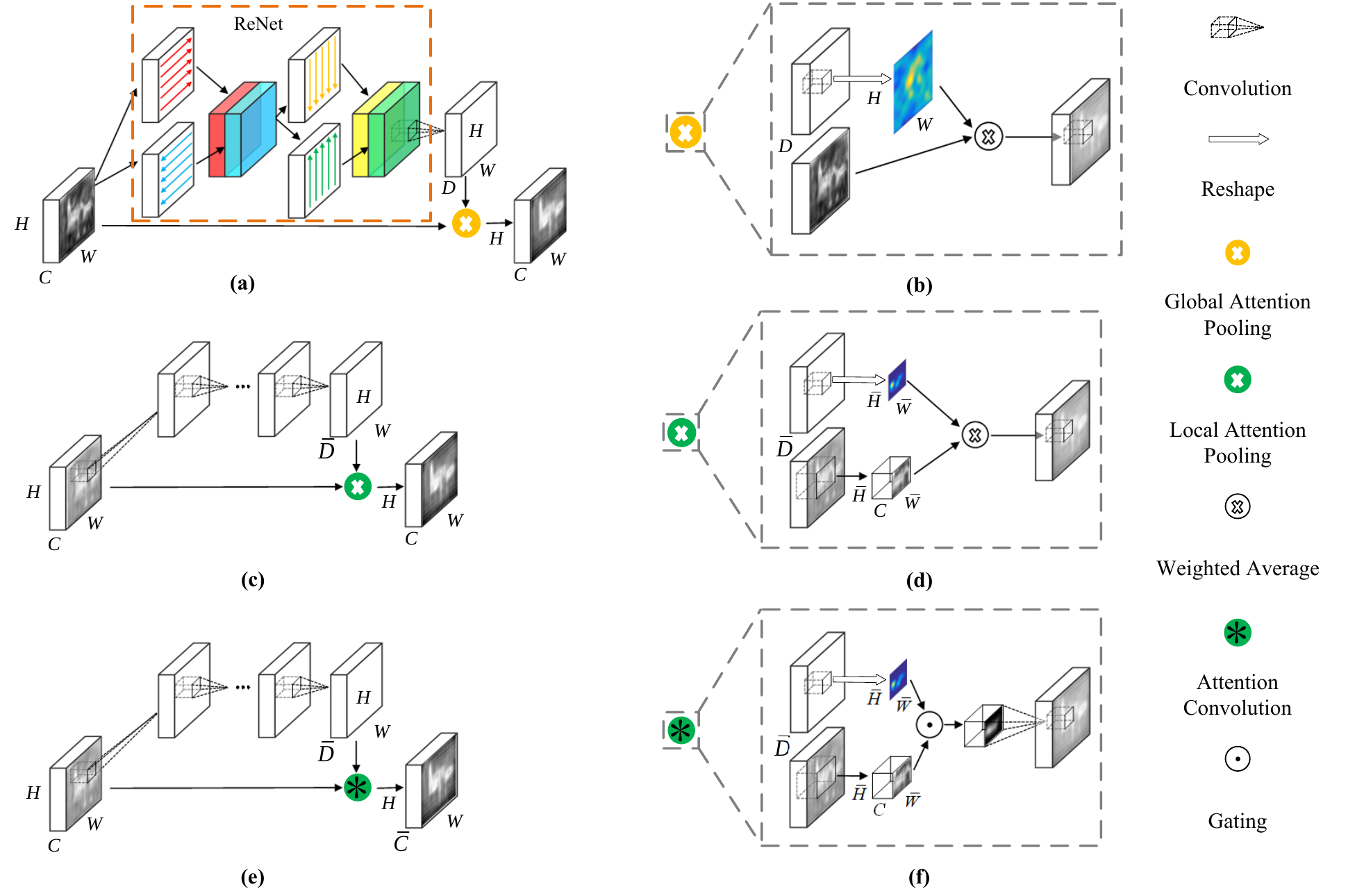}
  %(a)
  \put(5.8,53.5){\scriptsize $\bm{F}$}
  \put(39.5,53.7){\scriptsize $\bm{F}_{GAP}$}
  \put(35,59.9){\scriptsize $\bm{\alpha}$}
  %(b)
  \put(62.5,63){\scriptsize $\bm{\alpha}$}
  \put(68.7,63){\scriptsize $\bm{\alpha}^{w,h}$}
  \put(62.5,55.5){\scriptsize $\bm{F}$}
  \put(78.2,60.5){\scriptsize $\bm{F}_{GAP}$}
  %(c)
  \put(6,34){\scriptsize $\bm{F}$}
  \put(32.5,34.2){\scriptsize $\bm{F}_{LAP}$}
  \put(27,40.8){\scriptsize $\bar{\bm{\alpha}}$}
  %(d)
  \put(62.2,40.5){\scriptsize $\bar{\bm{\alpha}}$}
  \put(66.7,38.7){\scriptsize $\bar{\bm{\alpha}}^{w,h}$}
  \put(62.2,33.3){\scriptsize $\bm{F}$}
  \put(64.6,32.3){\scriptsize $\bar{\bm{F}}^{w,h}$}
  \put(76.9,38.3){\scriptsize $\bm{F}_{LAP}$}
  %(e)
  \put(6,12){\scriptsize $\bm{F}$}
  \put(32.5,12.2){\scriptsize $\bm{F}_{AC}$}
  \put(27,19){\scriptsize $\bm{g}$}
  %(f)
  \put(62.2,18.8){\scriptsize $\bm{g}$}
  \put(66.8,16.9){\scriptsize $\bm{g}^{w,h}$}
  \put(62.2,11.6){\scriptsize $\bm{F}$}
  \put(64,10.3){\scriptsize $\bar{\bm{F}}^{w,h}$}
  \put(77.7,17.1){\scriptsize $\bm{F}_{AC}$}
  \end{overpic}
  \caption{\textbf{Illustration of the proposed PiCANets.} (a)(c)(e) illustrate the proposed global attention pooling, local attention pooling, and attention convolution network architectures, respectively. (b)(d)(f) show detailed operations of GAP, LAP, and AC, respectively.}
  \label{PiCANetFig}
  \vspace{-0.3cm}
\end{figure*}
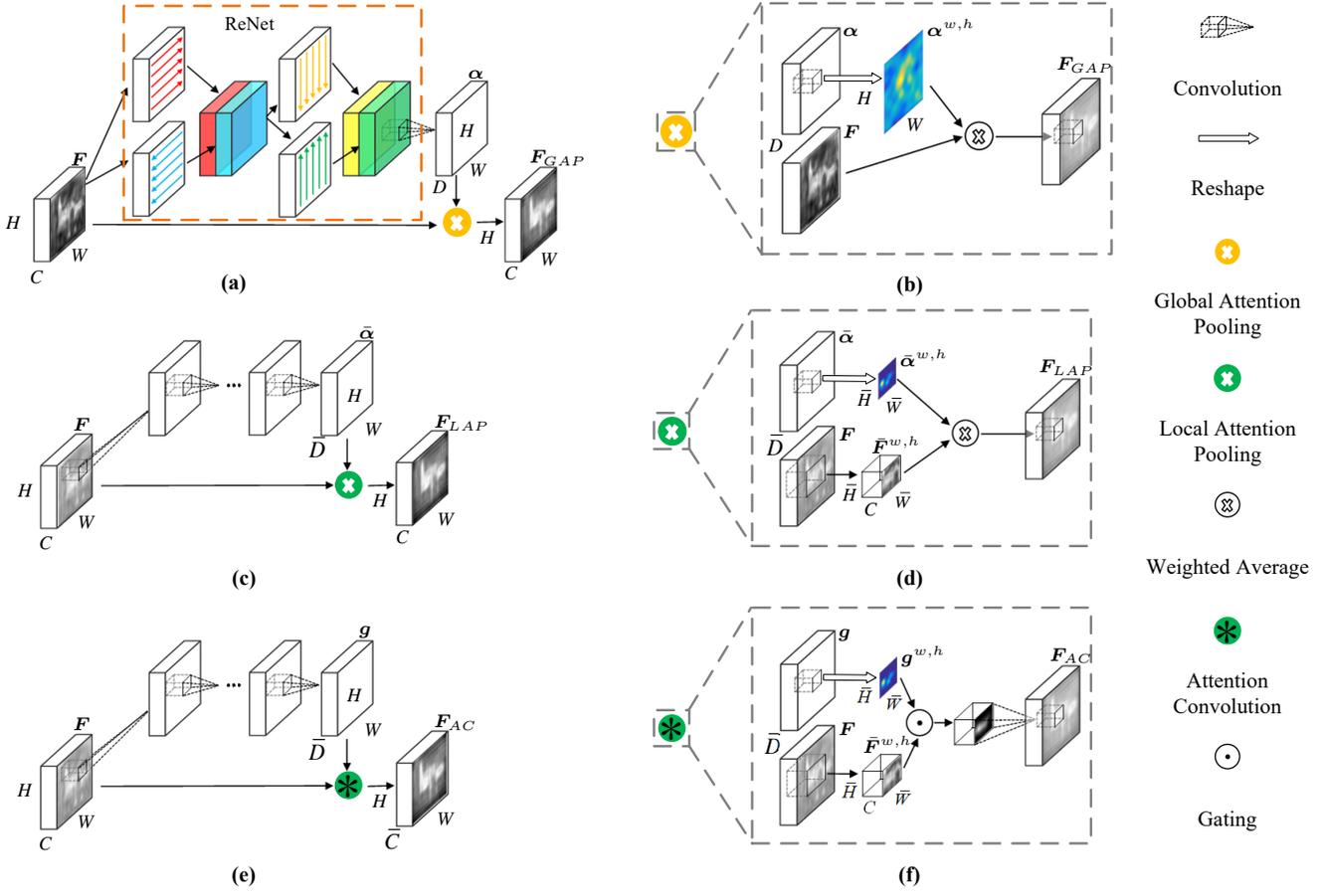

In this section, we give detailed formulations of the proposed three forms of the PiCANet. Suppose we have a convolutional (Conv) feature map $\bm{F}\in{\mathbb{R}^{W\times{H\times C}}}$, with $W$, $H$, and $C$ denoting its width, height and number of channels, respectively. For each location $(w,h)$ in $\bm{F}$, the GAP module generates global attention over the entire feature map $\bm{F}$, while the LAP module and the AC module generate attention over a local neighbouring region $\bar{\bm{F}}^{w,h}$ centered at $(w,h)$. As for the attention mechanism, GAP and LAP first adopt softmax to generate normalized attention weights and then use weighted average to pool the feature of each context location, while AC generates sigmoid attention weights and then uses them as gates to control the information flow of each context location in convolution.

%One generates attention over the whole input image, making every pixel to be able to selectively incorporate the global context. While the other ones generate each pixel an attention map over a neighboring local region to integrate informative local contextual information. As for how to adopt the attention weights to construct the attentive contextual feature for each pixel, we first use the traditional weighted average mechanism to aggregate the feature of every context location. Since this operation is similar to average pooling in CNNs, we refer the PiCANets using this mechanism with global and local contexts as \emph{global attention pooling} and \emph{local attention pooling}, respectively. For the local PiCANet, we can also integrate the attention mechanism with the convolution operation, where we use the attention weights as gates to control the information flow at each context location during conducting convolution for each pixel. We refer this form of the PiCANet as \emph{attention convolution}. All of the three kinds of PiCANet are fully differentiable and can be flexibly embedded into CNNs with end-to-end training.

\subsection{Global Attention Pooling}

The network architecture of GAP is shown in Figure~\ref{PiCANetFig}(a). To make all pixels have the capability of generating their own global attention in GAP, we first need them to be capable to ``see'' the whole feature map $\bm{F}$, where a network with the entire image as its receptive field is required. Although a fully connected layer is a straightforward choice, it has too many parameters to learn. Alternatively, we employ the ReNet model \cite{visin2015renet} as a more efficient and effective choice to perceive the global context, as shown in the orange dashed box in Figure~\ref{PiCANetFig}(a). To be specific, ReNet first deploys two LSTMs \cite{hochreiter1997lstm} along each row of $\bm{F}$, scanning the pixels one-by-one from left to right and from right to left, respectively. Then the two feature maps are concatenated to combine both left and right contextual information of each pixel. Next, ReNet uses another two LSTMs to scan each column of the obtained feature map in both bottom-up and up-bottom orders. Once again, the two feature maps are concatenated to combine both bottom and top contexts. By successively using horizontal and vertical bidirectional LSTMs to scan the feature maps, each pixel can remember its contextual information from all four directions, thus effectively integrating the global context. At the same time, ReNet can run each bidirectional LSTM in parallel and share the parameters for each pixel, thus is very efficient.

Following the ReNet module, a vanilla $1\times 1$ Conv layer is subsequently used to transform the output feature map to $D$ channels, where typically $D=W\times H$. Then we use the softmax activation function to obtain the normalized attention weights $\bm{\alpha}\in{\mathbb{R}^{W\times{H\times D}}}$. Specifically, at a pixel $(w,h)$ with the transformed feature $\bm{x}^{w,h}$, its attention weights $\bm{\alpha}^{w,h}$ can be obtained by:
\begin{equation} \label{attSoftmax}
\alpha_{i}^{w,h}=\frac{\exp{(x_{i}^{w,h})}}{\sum_{j=1}^D\exp{(x_{j}^{w,h})}},
\end{equation}
where $i\in{\{1,\ldots,D\}}$, and $\bm{x}^{w,h},\bm{\alpha}^{w,h}\in{\mathbb{R}^D}$. Concretely, $\alpha_{i}^{w,h}$ represents the attention weight of the $i^{th}$ location in $\bm{F}$ w.r.t. the referred pixel $(w,h)$.

Finally, as shown in Figure~\ref{PiCANetFig}(b), for each pixel, we utilize its attention weights to pool the features in $\bm{F}$ via weighted average. As a result, we obtain an attentive contextual feature map $\bm{F}_{GAP}$, at each location of which we have:
\begin{equation} \label{GAP}
\bm{F}_{GAP}^{w,h}=\sum_{i=1}^D\alpha_{i}^{w,h}\bm{f}_{i},
\end{equation}
where $\bm{f}_{i}\in{\mathbb{R}^C}$ is the feature at the $i^{th}$ location of $\bm{F}$, and $\bm{F}_{GAP}\in{\mathbb{R}^{W\times{H\times C}}}$. This operation is similar to the traditionally used pooling layer in CNNs, except that we adopt the generated attention weights to adaptively pool the features for context selection instead of using fixed pooling templates in each pooling window.

%-------------------------------------------------------------------------
\subsection{Local Attention Pooling}

The LAP module is similar to GAP except that it only operates over a local neighbouring region for each pixel, as shown in Figure~\ref{PiCANetFig}(c). To be specific, given the width $\bar{W}$ and the height $\bar{H}$ of the local region, we first deploy several vanilla Conv layers on top of $\bm{F}$ with their receptive field size achieving $\bar{W}\times\bar{H}$. Thus, we make each pixel $(w,h)$ be able to ``see'' the local neighbouring region $\bar{\bm{F}}^{w,h}\in{\mathbb{R}^{\bar{W}\times{\bar{H}\times C}}}$ centered at it. Then, similar to GAP, we use another Conv layer with $\bar{D}=\bar{W}\times\bar{H}$ channels and the softmax activation function to obtain the local attention weights $\bar{\bm{\alpha}}\in{\mathbb{R}^{W\times{H\times\bar{D}}}}$. At last, as shown in Figure~\ref{PiCANetFig}(d), for each pixel $(w,h)$, we use its attention weights $\bar{\bm{\alpha}}^{w,h}$ to obtain the attentive contextual feature $\bm{F}_{LAP}^{w,h}$ as the weighted average of $\bar{\bm{F}}^{w,h}$:
\begin{equation} \label{LAP}
\bm{F}_{LAP}^{w,h}=\sum_{i=1}^{\bar{D}}{{\bar{\alpha}}_i^{w,h}\bar{\bm{f}}_{i}^{w,h}},
\end{equation}
where $\bar{\bm{f}}_{i}^{w,h}$ is the feature at the $i^{th}$ location of $\bar{\bm{F}}^{w,h}$, and $\bm{F}_{LAP}\in{\mathbb{R}^{W\times{H\times C}}}$.

%-------------------------------------------------------------------------
\subsection{Attention Convolution}

Similar to LAP, the proposed AC module also generates and utilizes local attention for each pixel. The difference is that AC generates sigmoid attention weights and adopts them as gates to control whether each context location needs to be involved in the convolutional feature extraction for the center pixel. The detailed network architecture is shown in Figure~\ref{PiCANetFig}(e). Given the Conv kernel size $\bar{W}\times\bar{H}$ and the number of output channels $\bar{C}$, similar Conv layers are first used as in LAP to generate local attention gates $\bm{g}\in{\mathbb{R}^{W\times{H\times\bar{D}}}}$ except that in AC we use the sigmoid activation function for the last Conv layer. Following \eqref{attSoftmax}, we have:
\begin{equation} \label{attSigmoid}
g_{i}^{w,h}=\frac{1}{1+\exp{(-x_{i}^{w,h})}},
\end{equation}
where $i\in{\{1,\ldots,\bar{D}\}}$, and $g_{i}^{w,h}$ is the attention gate of the $i^{th}$ location in $\bar{\bm{F}}^{w,h}$, determining whether its information should flow to the next layer for the feature extraction at $(w,h)$.

Subsequently, we adopt $\bm{g}$ into a convolution layer on top of $\bm{F}$, where the detailed operations are shown in Figure~\ref{PiCANetFig}(f). To be specific, at pixel $(w,h)$, we first use the attention gates $\bm{g}^{w,h}$ to modulate the features in $\bar{\bm{F}}^{w,h}$ via pixel-wise multiplication, then we multiply the result feature matrix with the convolution weight matrix $\bm{W}\in{\mathbb{R}^{\bar{W}\times{\bar{H}\times{C\times{\bar{C}}}}}}$ to obtain the attentive contextual feature $\bm{F}_{AC}^{w,h}$. By decomposing convolution into per-location operation, we have:
\begin{equation} \label{AC}
\bm{F}_{AC}^{w,h}=\sum_{i=1}^{\bar{D}}{g_i^{w,h}\bar{\bm{f}}_{i}^{w,h}\bm{W}_{i}}+\bm{b},
\end{equation}
where $\bm{W}_{i}\in{\mathbb{R}^{C\times{\bar{C}}}}$ is the $i^{th}$ spatial element of $\bm{W}$, and $\bm{b}\in{\mathbb{R}^{\bar{C}}}$ is the convolution bias. The obtained attentive feature map $\bm{F}_{AC}\in{\mathbb{R}^{W\times{H\times{\bar{C}}}}}$.

Compared to LAP, AC introduces further non-linear transformation on top of the attended features, which may lead to more discriminative feature abstraction but with more parameters to learn.

%-------------------------------------------------------------------------
\subsection{Effective and Efficient Implementation}

The per-pixel attending operation of the proposed PiCANets can be easily conducted in parallel for all pixels via GPU acceleration. The dilation convolution algorithm \cite{chen2018deeplab} can also be adopted to uniformly-spaced sample distant context locations in the attending process of each pixel. In this way, we can efficiently attend to large context regions with significantly reduced computational cost by using a small $D$ or $\bar{D}$ with dilation. Meanwhile, all the three formulations \eqref{GAP}\eqref{LAP}\eqref{AC} of the PiCANets are fully differentiable, thus enable end-to-end training with other Convnet modules via the back-propagation algorithm \cite{rumelhart1988bp}. When using deep layers to generate the attention weights, batch normalization (BN) \cite{ioffe2015bn} can also be used to facilitate the gradient propagation, making the attention learning more effective.

%-------------------------------------------------------------------------
\subsection{Difference with Prior Work}

In \cite{vaswani2017attention}, Vaswani \emph{et al}. also propose an attention model for machine translation where each word in the input or the output sequence can attend to its corresponding global or local context positions. Our work differs with theirs in several aspects. First, their model embeds the attention modules into feedforward networks for machine translation while our model adopts them in CNNs for saliency detection and other dense prediction vision tasks. Second, their model generates attention over 1D word positions while our model extends the attention mechanism to 2D spatial locations. Third, their attention is generated by the dot-product between the query and the keys while we use ReNet or Conv layers to learn to generate the attention weights automatically. Forth, we also propose the attention convolution operation to introduce effective non-linear transformation for the attentive feature extraction.

Dauphin \etal \cite{dauphin2016language} also present a gated convolution network for language modeling. However, their model is also proposed for 1D convolution while ours performs for 2D spatial convolution. Furthermore, their attention gates are actually applied on the output channels, while ours work on the input spatial locations, which is totally different.

%------------------------------------------------------------------------
\section{Salient Object Detection with PiCANets}

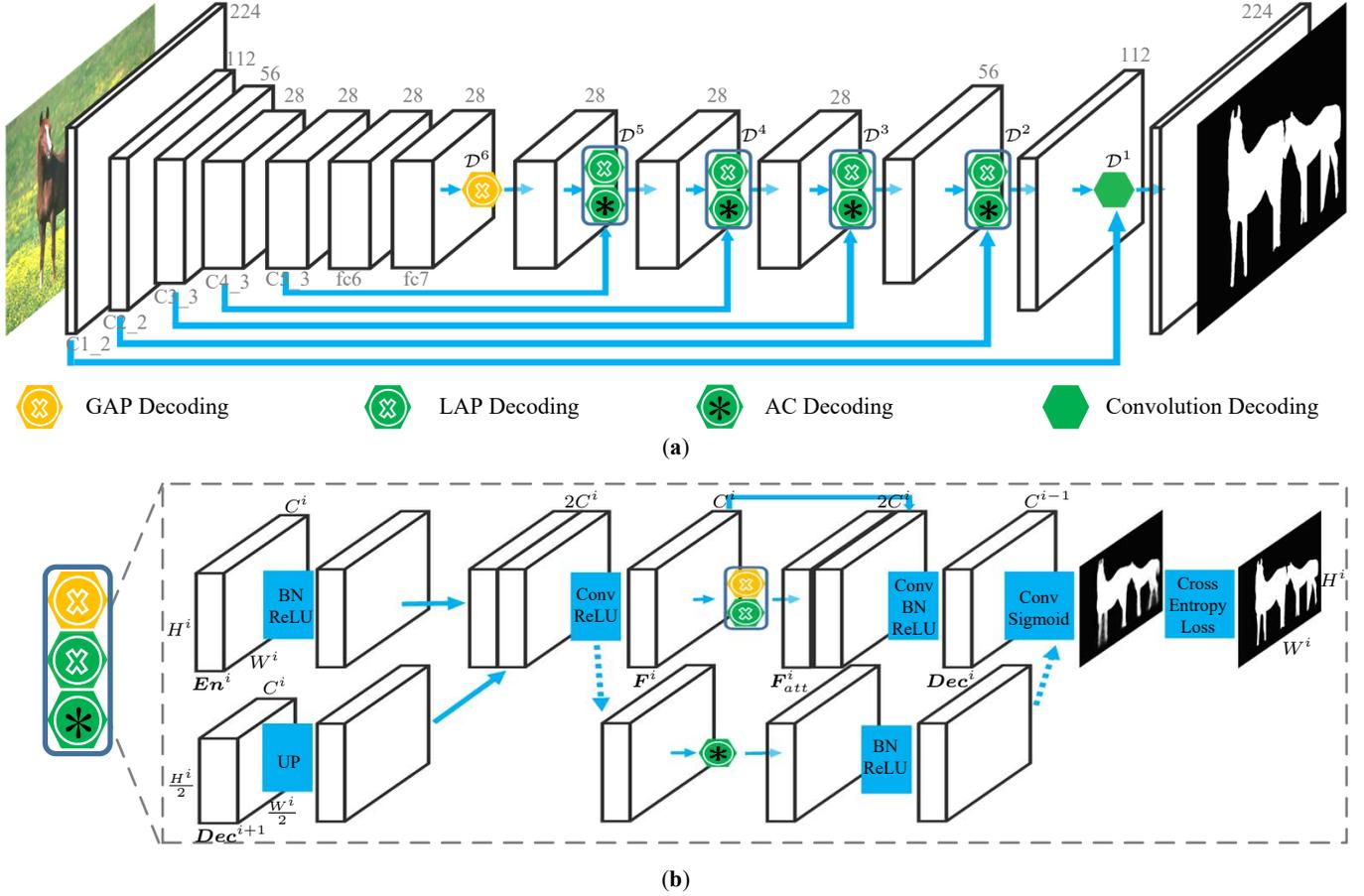
\begin{figure*}[!t]
  %\graphicspath{{Figures/PiCANet/}}
  \centering
  %\includegraphics[width=1\linewidth]{SO_net.pdf}
  %\begin{overpic}[width=1\linewidth,grid,tics=1]{SO_net.pdf}
  \begin{overpic}[width=1\linewidth]{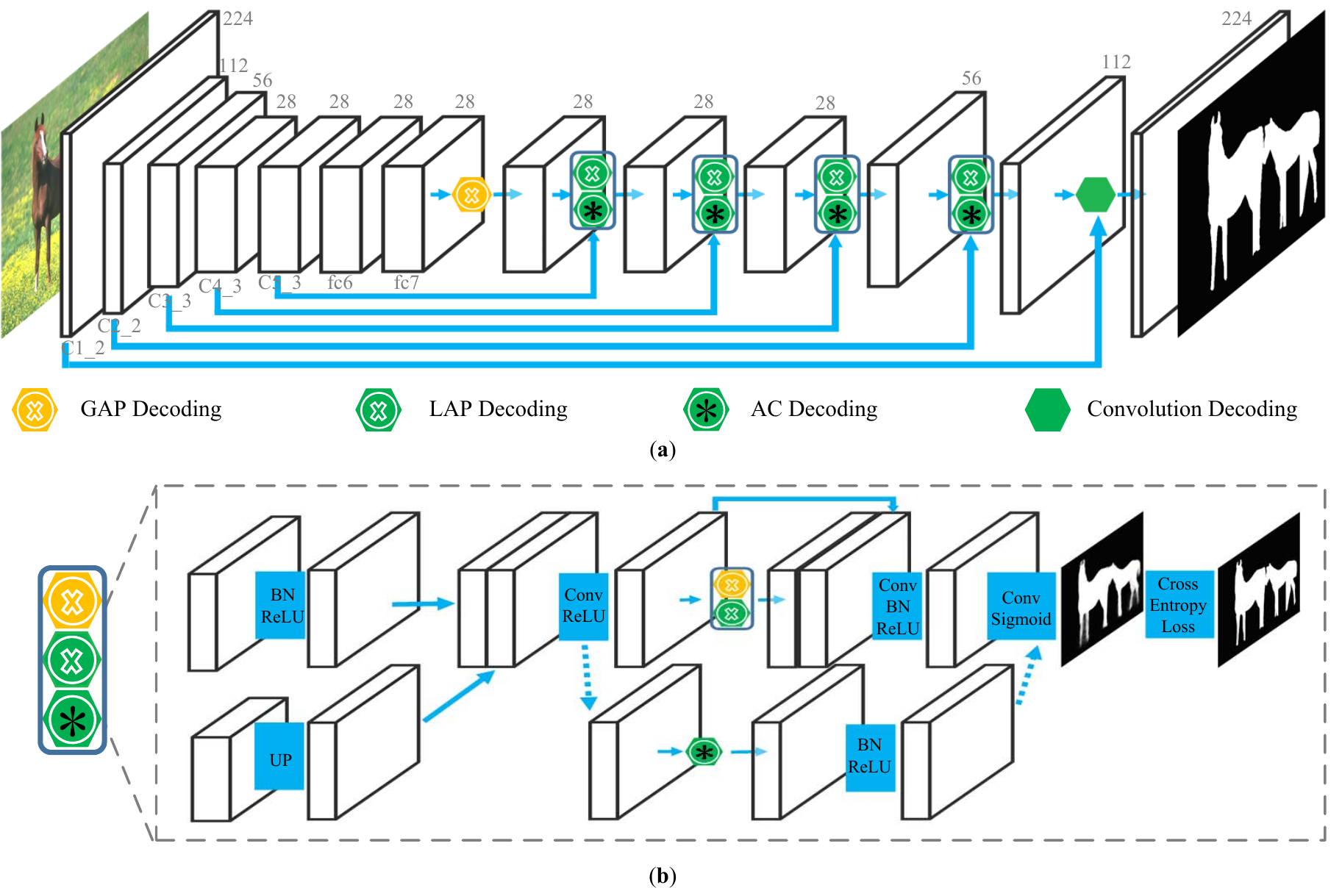}
  \put(33.8,53.9){\scriptsize $\mathcal D^6$}
  \put(45.2,55.8){\scriptsize $\mathcal D^5$}
  \put(54.2,55.8){\scriptsize $\mathcal D^4$}
  \put(63.2,55.8){\scriptsize $\mathcal D^3$}
  \put(73.6,55.8){\scriptsize $\mathcal D^2$}
  \put(81,53.9){\scriptsize $\mathcal D^1$}
  \put(19,15.1){\scriptsize $C^i$}
  \put(11.9,8){\scriptsize $\frac{H^i}{2}$}
  \put(19.2,6){\scriptsize $\frac{W^i}{2}$}
  \put(20.6,28.5){\scriptsize $C^i$}
  \put(11.9,19.4){\scriptsize $H^i$}
  \put(18,17){\scriptsize $W^i$}
  \put(41.2,28.8){\scriptsize $2C^i$}
  \put(52.1,28.8){\scriptsize $C^i$}
  \put(64.2,28.8){\scriptsize $2C^i$}
  \put(75,28.8){\scriptsize $C^{i-1}$}
  \put(14,4.2){\scriptsize $\bm{Dec}^{i+1}$}
  \put(13.8,15.4){\scriptsize $\bm{En}^i$}
  \put(46.2,15.6){\scriptsize $\bm{F}^i$}
  \put(56.3,15.6){\scriptsize $\bm{F}_{att}^i$}
  \put(67.8,15.6){\scriptsize $\bm{Dec}^i$}
  \put(96.8,23){\scriptsize $H^i$}
  \put(94,18){\scriptsize $W^i$}
  \end{overpic}
  \caption{\textbf{Architecture of the proposed saliency network with PiCANets.} (a) Overall architecture of our saliency network. For simplicity, we only show the last layer of each block in the VGG network, \ie the C*\_* layers and fc* layers. We use $\mathcal D^i$ to indicate the $i^{th}$ decoding module. The spatial sizes are marked over the cuboids which represent the feature maps. (b) Illustration of an attentive decoding module, either using GAP, LAP, or AC. We use $\bm{En}^i$ and $\bm{Dec}^i$ to denote the $i^{th}$ encoding feature map or decoding feature map, respectively. While $\bm{F}^i$ and $\bm{F}_{att}^i$ are used to denote the $i^{th}$ fusion feature map and the attentive contextual feature map, respectively. ``UP'' denotes upsampling. Some crucial spatial sizes and channel numbers are also marked. Since using AC leads to a slightly different network structure compared with using GAP and LAP, we use dashed arrows to denote the different part of the network path.}
  \label{SONetFig}
  \vspace{-0.3cm}
\end{figure*}

In this section, we elaborate on how we hierarchically adopt PiCANets for incorporating multiscale attentive contexts to detect salient objects. The whole network is based on a U-Net \cite{ronneberger2015unet} architecture as shown in Figure~\ref{SONetFig}(a). Different from \cite{ronneberger2015unet}, we use dilation convolution \cite{chen2018deeplab} in the encoder network to keep large sizes of the feature maps for avoiding losing too many spatial details. The decoder follows the idea of U-Net to use skip connections but particularly with our proposed PiCANets embedded.

%-------------------------------------------------------------------------
\subsection{Encoder Network}

Considering the GAP module requires the input feature map to have a fixed size, we directly resize images to a fixed size of $224\times 224$ as the network input. The encoder part of our model is an FCN with a pretrained backbone network, as which we select the VGG \cite{simonyan2014vgg} 16-layer network for a fair comparison. The VGG-16 net contains 13 Conv layers, 5 max-pooling layers, and 2 fully connected layers. As shown in Figure~\ref{SONetFig}(a), in order to preserve relative large spatial sizes in higher layers for accurate saliency detection, we reduce the pooling strides of the pool4 and pool5 layers to be 1 and introduce dilation of 2 for the Conv kernels in the Conv5 block. We also follow \cite{chen2018deeplab} to transform the last 2 fully connected layers to Conv layers for preserving the rich high-level features learned in them. To be specific, we set the fc6 layer to have 1024 channels and $3\times 3$ Conv kernels with dilation of 12 while the fc7 layer is set to have the same channel number with $1\times 1$ Conv kernels. Thus, the stride of the whole encoder network is reduced to 8, and the spatial size of the final feature map is $28\times 28$.

%-------------------------------------------------------------------------
\subsection{Decoder Network}\label{sec:decoder}

Next, we elaborate our decoder part. As shown in Figure~\ref{SONetFig}(a), the decoder network has six decoding modules, named $\mathcal D^6,\mathcal D^5,\ldots,\mathcal D^1$ in sequential order. In $\mathcal D^i$, usually we generate a decoding feature map $\bm{Dec}^i$ by fusing the preceding decoding feature $\bm{Dec}^{i+1}$ with an intermediate encoder feature map $\bm{En}^i$. We select $\bm{En}^i$ as the last Conv feature map before the ReLU activation of the $i^{th}$ Conv block in the encoder part, where its size is denoted as $W^i\times{H^i\times C^i}$ and all the six selected encoder feature maps are marked in Figure~\ref{SONetFig}(a). An exception is that in $\mathcal D^6$, $\bm{Dec}^6$ is directly generated from $\bm{En}^6$ without the preceding decoding feature map and $\bm{En}^6$ comes from the fc7 layer.

The detailed decoding process is shown in Figure~\ref{SONetFig}(b). Specifically, we first pass $\bm{En}^i$ through a BN layer and the ReLU activation for normalization and non-linear transformation to get ready for the subsequent fusion. As for $\bm{Dec}^{i+1}$, usually it has a half size of $W^i/2\times{H^i/2}$, thus we upsample it to $W^i\times H^i$ via bilinear interpolation. Next, we concatenate $\bm{En}^i$ with the upsampled $\bm{Dec}^{i+1}$ and fuse them into a feature map $\bm{F}^i$ with $C^i$ channels by using a Conv layer and the ReLU activation. Then we utilize either GAP, LAP, or AC on $\bm{F}^i$ to obtain the attentive contextual feature map $\bm{F}_{att}^i$, where we use $\bm{F}_{att}$ as the general denotation of $\bm{F}_{GAP}$, $\bm{F}_{LAP}$, and $\bm{F}_{AC}$. Since for GAP and LAP, at each pixel $\bm{F}_{att}^i$ is simply a linear combination of $\bm{F}^i$, we use it as complementary information for the original feature. Thus we concatenate and fuse $\bm{F}^i$ and $\bm{F}_{att}^i$ into $\bm{Dec}^i$ via a Conv layer with BN and the ReLU activation. We keep the spatial size of $\bm{Dec}^i$ as $W^i\times H^i$ but set its number of channels to be the same as that of $\bm{En}^{i-1}$, \ie $C^{i-1}$. For AC, as it has already merged the attention and convolution operations, we directly set its number of output channels to be $C^{i-1}$ and generate $\bm{Dec}^i$ after using BN and the ReLU activation, which is shown as the dashed path in Figure~\ref{SONetFig}(b).

Since GAP conducts the attention operation over the whole feature map, which is computationally costly, we only use it in early decoding modules that have small feature maps but with high-level semantics. Finally, we find that adopting GAP in $\mathcal D^6$ and using LAP or AC in latter modules leads to the best performance. For computational efficiency, we do not use any PiCANet in $\mathcal D^1$, in which case $\bm{En}^1$ and $\bm{Dec}^2$ are directly fused into $\bm{Dec}^1$ by vanilla Conv layers. Analyses of the network settings with different usage of PiCANets can be found in Section~\ref{sec:ablation}.

%-------------------------------------------------------------------------
\subsection{Training Loss}

\begin{table*} [!ht]
\begin{center}
\caption{\textbf{Quantitative comparison of different model settings for saliency detection.} ``*GAP'', ``*AC'', and ``*LAP'' mean we embed these PiCANets in corresponding decoding modules. ``LC'', ``MaxP'', and ``AveP'' mean large-kernel convolution, max-pooling, and average pooling, respectively. \red{Red} indicates the best performance.}
\vspace{-0.2cm}
\label{ablationTab}
\footnotesize
\begin{tabular}{@{}L{2cm}|C{0.45cm}C{0.45cm}C{0.45cm}|C{0.45cm}C{0.45cm}C{0.45cm}|C{0.45cm}C{0.45cm}C{0.45cm}|C{0.45cm}C{0.45cm}C{0.45cm}|C{0.45cm}C{0.45cm}C{0.45cm}|C{0.45cm}C{0.45cm}C{0.45cm}}
\toprule
Dataset & \multicolumn{3}{c|}{SOD \cite{yang2013gbmr}} & \multicolumn{3}{c|}{ECSSD \cite{yang2013gbmr}} & \multicolumn{3}{c|}{PASCAL-S \cite{li2014secrets}} & \multicolumn{3}{c|}{HKU-IS \cite{li2015mdf}} & \multicolumn{3}{c|}{DUT-O \cite{yang2013gbmr}} & \multicolumn{3}{c}{DUTS-TE \cite{wang2017duts}} \\ \cmidrule{1-19}
Metric
&$F_{\beta}$&  $S_m$    &   MAE     &$F_{\beta}$&  $S_m$    &   MAE     &$F_{\beta}$&  $S_m$    &   MAE     &$F_{\beta}$&  $S_m$    &   MAE     &$F_{\beta}$&  $S_m$    &   MAE     &$F_{\beta}$&  $S_m$    &   MAE     \\ \midrule
\multicolumn{19}{c}{Baseline}\\ \midrule
U-Net \cite{ronneberger2015unet}
&0.836      &0.753      &0.122      &0.906      &0.886      &0.052      &0.852      &0.809      &0.097      &0.894      &0.877      &0.045      &0.762      &0.794      &0.072      &0.823      &0.834      &0.057
\\ \midrule
\multicolumn{19}{c}{Progressively embedding PiCANets}\\ \midrule
+6GAP
&0.839      &0.759      &0.119      &0.915      &0.896      &0.049      &0.862      &0.818      &0.094      &0.903      &0.887      &0.044      &0.784      &0.810      &0.070      &0.837      &0.845      &0.056
\\
+6GAP\_5AC
&0.847      &0.773      &0.114      &0.921      &0.903      &0.048      &0.868      &0.826      &0.091      &0.910      &0.894      &0.043      &0.786      &0.817      &0.069      &0.843      &0.852      &0.054
\\
+6GAP\_54AC
&0.853      &0.780      &0.110      &0.927      &0.910      &0.045      &0.872      &0.829      &0.090      &0.915      &0.901      &0.041      &0.797      &0.825      &0.067      &0.851      &0.858      &0.054
\\
+6GAP\_543AC
&0.863      &0.789      &\red{0.105}&0.933      &0.915      &0.045      &0.877      &0.832      &0.089      &0.921      &0.906      &0.040      &0.803      &0.830      &0.068      &0.854      &0.862      &0.054
\\
+6GAP\_5432AC
&0.858      &0.786      &0.107      &0.935      &\red{0.917}&\red{0.044}&\red{0.883}&\red{0.838}&\red{0.085}&\red{0.924}&\red{0.908}&\red{0.039}&\red{0.808}&\red{0.835}&0.065      &\red{0.859}&\red{0.867}&\red{0.051}
\\ \midrule
\multicolumn{19}{c}{Different embedding settings}\\ \midrule
+65432AC
&0.858      &0.784      &0.110      &0.932      &0.914      &0.045      &0.880      &0.831      &0.087      &0.922      &0.904      &0.040      &0.805      &0.831      &\red{0.063}&\red{0.859}&0.865      &\red{0.051}
\\
+65GAP\_432AC
&\red{0.866}&\red{0.795}&\red{0.105}&\red{0.937}&\red{0.917}&\red{0.044}&0.876      &0.835      &0.088      &\red{0.924}&\red{0.908}&\red{0.039}&0.805      &0.832      &0.067      &0.858      &0.866      &0.052
\\
+654GAP\_32AC
&0.859      &0.785      &0.107      &0.935      &0.916      &\red{0.044}&0.877      &0.835      &0.087      &0.922      &0.905      &0.040      &0.802      &0.828      &0.066      &0.855      &0.864      &0.052
\\ \midrule
\multicolumn{19}{c}{AC vs. LAP?}\\ \midrule
+6GAP\_5432LAP
&\red{0.866}&0.788      &0.106      &0.934      &0.916      &\red{0.044}&0.880      &0.835      &0.087      &0.923      &0.905      &0.040      &0.799      &0.829      &0.066      &0.857      &0.862      &0.052
\\ \midrule
\multicolumn{19}{c}{Attention loss?}\\ \midrule{}
+6GAP\_5432AC
&\mlr{0.857}&\mlr{0.785}&\mlr{0.109}&\mlr{0.930}&\mlr{0.913}&\mlr{0.045}&\mlr{0.879}&\mlr{0.835}&\mlr{0.086}&\mlr{0.921}&\mlr{0.905}&\mlr{0.040}&\mlr{0.801}&\mlr{0.829}&\mlr{0.066} &\mlr{0.856}&\mlr{0.863}&\mlr{0.053} \\
\_w/o\_$\bm{L}_{GA}^6$
&           &           &           &           &           &           &           &           &           &           &           &           &           &           &           &           &           &
\\ \midrule
\multicolumn{19}{c}{Comparison with vanilla pooling and Conv layers}\\ \midrule
+6ReNet\_5432LC
&0.851      &0.774      &0.114      &0.920      &0.900      &0.049      &0.866      &0.820      &0.093      &0.907      &0.891      &0.043      &0.786      &0.816      &0.071      &0.841      &0.850      &0.056
\\
+6G\_5432L\_AveP
&0.842      &0.770      &0.114      &0.918      &0.899      &0.048      &0.865      &0.823      &0.092      &0.905      &0.889      &0.043      &0.782      &0.811      &0.071      &0.837      &0.847      &0.056
\\
+6G\_5432L\_MaxP
&0.845      &0.771      &0.116      &0.918      &0.899      &0.048      &0.866      &0.819      &0.093      &0.905      &0.889      &0.042      &0.776      &0.808      &0.070      &0.838      &0.848      &0.055
\\
\bottomrule
\end{tabular}
\vspace{-0.3cm}
\end{center}{}
\end{table*}

To facilitate the network training, we adopt deep supervision for each decoding module. Specifically, in $\mathcal D^i$, we use a Conv layer with one output channel and the sigmoid activation on top of $\bm{Dec}^i$ to generate a saliency map $\bm{S}^i$ with size $W^i\times H^i$. Then, the ground truth saliency map is resized to the same size, which is denoted as $\bm{G}^i$, to supervise the network training based on the average cross-entropy saliency loss $\bm{L}_S^i$:
\begin{equation} \label{SaliencyLoss}
\begin{split}
\bm{L}_S^i=&-\frac{1}{W^iH^i}\sum_{w=1}^{W^i}\sum_{h=1}^{H^i}\bm{G}^i(w,h)\log\bm{S}^i(w,h)\\
&+(1-\bm{G}^i(w,h))\log(1-\bm{S}^i(w,h)),
\end{split}
\end{equation}
where $\bm{G}^i(w,h)$ and $\bm{S}^i(w,h)$ denote their saliency values at the location $(w,h)$.

In our preliminary version of this work \cite{liu2018picanet}, the global attention is found to be able to learn global contrast, \ie the attention map of foreground pixels mainly highlights background regions and vice verse. However, the global attention maps are usually inaccurate and dispersive as being implicitly learned. Thus, we also propose to explicitly learn the global attention in GAP. Specifically, we simulate the global contrast mechanism to extract foreground and background regions from the ground truth saliency maps for supervising the learning of the global attention at background and foreground pixels, respectively. We take $\mathcal D^6$ as the example. First, we generate the normalized ground truth global attention map $\bm{A}^{w,h}$ for each pixel $(w,h)$ in $\bm{F}^6$:
\begin{equation} \label{GroundTruthAttention}
\bm{A}^{w,h}=
\begin{cases}
\dfrac{\bm{G}^6}{\sum\bm{G}^6},     & \text{if}\ \bm{G}^6(w,h)=0,\\
\dfrac{1-\bm{G}^6}{\sum(1-\bm{G}^6)}, & \text{if}\ \bm{G}^6(w,h)=1.
\end{cases}
\end{equation}
Then, we use the averaged KL divergence loss between $\bm{A}^{w,h}$ and $\bm{\alpha}^{w,h}$ at each pixel as the global attention loss $\bm{L}_{GA}^6$:
\begin{equation} \label{GALoss}
\bm{L}_{GA}^6=\frac{1}{W^6H^6}\sum_{w,w'=1}^{W^6}\sum_{h,h'=1}^{H^6}\bm{A}^{w,h}(w',h')\log\frac{\bm{A}^{w,h}(w',h')}{\bm{\alpha}^{w,h}(w',h')},
\end{equation}
where $\bm{\alpha}^{w,h}(w',h')=\alpha_{(h'-1)W^6+w'}^{w,h}$.

At last, the final loss is obtained by a weighted sum of the saliency losses in different decoding modules and the global attention loss:
\begin{equation} \label{Loss}
\bm{L}=\sum_{i=1}^{6}\gamma^i\bm{L}_S^i+\gamma^{GA}\bm{L}_{GA}^6.
\end{equation}

%------------------------------------------------------------------------
\section{Experiments}
In this section, we evaluate the effectiveness of the proposed PiCANets and the saliency model via substantial experiments on six saliency benchmark datasets. Furthermore, we also validate that PiCANets can benefit other general dense prediction tasks, \eg, semantic segmentation, and object detection.

%-------------------------------------------------------------------------
\subsection{Datasets}

We use six widely used saliency benchmark datasets to evaluate our method. \textbf{SOD} \cite{movahedi2010sod} contains 300 images with complex backgrounds and multiple foreground objects. \textbf{ECSSD} \cite{yan2013hs} has 1,000 semantically meaningful and complex images. The \textbf{PASCAL-S} \cite{li2014secrets} dataset consists of 850 images selected from the PASCAL VOC 2010 segmentation dataset. \textbf{DUT-O} \cite{yang2013gbmr} includes 5,168 challenging images, each of which usually has complicated background and one or two foreground objects. \textbf{HKU-IS} \cite{li2015mdf} contains 4,447 images with low color contrast and multiple foreground objects in each image. The last one is the \textbf{DUTS} \cite{wang2017duts} dataset, which is currently the largest salient object detection benchmark dataset. It contains 10,553 images in the training set, \ie DUTS-TR, and 5,019 images in the test set, \ie DUTS-TE. Most of these images have challenging scenarios for saliency detection.

%-------------------------------------------------------------------------
\subsection{Evaluation Metrics}

We adopt four evaluation metrics to evaluate our model. The first one is the precision-recall (PR) curve. Specifically, a predicted saliency map $\bm{S}$ is first binarized by a threshold and then compared with the corresponding ground truth saliency map $\bm{G}$. By varying the threshold between 0 to 255, we can obtain a series of precision-recall value pairs to draw the PR curve.

The second metric is the F-measure score which comprehensively considers both precision and recall:
\begin{equation} \label{fmeasure}
F_{\beta}=\frac{(1+\beta^2)Precision\times Recall}{\beta^2 Precision+Recall},
\end{equation}
where we set $\beta^2$ to 0.3 as suggested in previous work. Finally, we report the max F-measure score under the optimal threshold.

The third metric we use is the Mean Absolute Error (MAE). It computes the average absolute per-pixel difference between $\bm{S}$ and $\bm{G}$:
\begin{equation} \label{MAE}
MAE=\frac{1}{WH}\sum_{w=1}^W\sum_{h=1}^H\left|\bm{G}(w,h)-\bm{S}(w,h)\right|.
\end{equation}

All of the three above metrics are based on pixel-wise errors and seldom take structural knowledge into account. Thus we also follow \cite{zhang2018pagrn,fan2018SOC,hsu2018unsupervised} to adopt the Structure-measure \cite{fan2017structure} metric $S_m$ for evaluating both region-aware and object-aware structural similarities between $\bm{S}$ and $\bm{G}$. We use the same weight as \cite{fan2017structure} to take the average of the two kinds of similarities as the $S_m$ score.

%-------------------------------------------------------------------------
\subsection{Implementation Details}

\Paragraph{Network structure.}
In the decoding modules, all of the Conv kernels in Figure~\ref{SONetFig}(b) are set to $1\times 1$. In the GAP module, we use 256 hidden neurons for the ReNet, and then we use a $1\times 1$ Conv layer to generate $D=100$ dimensional attention weights, which can be reshaped to $10\times 10$ attention maps. In its attending operation, we use $dilation=3$ to attend to the $28\times 28$ global context. In each LAP module or AC module, we first use a $7\times 7$ Conv layer with $dilation=2$, zero padding, and the ReLU activation to generate an intermediate feature map with 128 channels. Then we adopt a $1\times 1$ Conv layer to generate $\bar{D}=49$ dimensional attention weights, from which $7\times 7$ attention maps can be obtained. Thus we can attend to $13\times 13$ local context regions with $dilation=2$ and zero padding.

\vspace{3mm}
\Paragraph{Training and testing.}
We follow \cite{Wang2017srm,zhang2018bmp,wang2018dgrl,zhang2018pagrn} to use the DUTS-TR set as our training set. For data augmentation, we simply resize each image to $256\times 256$ with random mirror-flipping and then randomly crop $224\times 224$ image regions for training. The whole network is trained end-to-end using stochastic gradient descent (SGD) with momentum. As for the weight of each loss term in \eqref{Loss}, we empirically set $\gamma^6,\gamma^5,\ldots,\gamma^1$ as 0.5, 0.5, 0.5, 0.8, 0.8, and 1, respectively, without further tuning. While $\gamma^{GA}$ is set to 0.2 based on the performance validation. We train the decoder part with random initialization and the learning rate of 0.01 and finetune the encoder with a 0.1 times smaller learning rate. We set the batchsize to 9, the maximum iteration step to 40,000, and use the ``multistep'' policy to decay the learning rates by a factor of 0.1 at the 20,000$th$ and the 30,000$th$ step. The momentum and the weight decay are set to 0.9 and 0.0005, respectively.

We implement our model based on the Caffe \cite{jia2014caffe} library. A GTX 1080 Ti GPU is used for acceleration. When testing, each image is directly resized to $224\times 224$ and fed into the network, then we can obtain its predicted saliency map from the network output without any post-processing. The prediction process only costs 0.127s for each image. Our code is available at \url{https://github.com/nian-liu/PiCANet}.

%-------------------------------------------------------------------------
\subsection{Ablation Study}\label{sec:ablation}

\Paragraph{Progressively embedding PiCANets.}
To demonstrate the effectiveness of progressively embedding the proposed PiCANets in the decoder network, we show quantitative comparison results of different model settings in Table~\ref{ablationTab}. We first take the basic U-Net \cite{ronneberger2015unet} as our baseline model and then progressively embed global and local PiCANets into the decoding modules as described in Section~\ref{sec:decoder}. For the local PiCANets, which include both LAP and AC, we take the latter as the example here. In Table~\ref{ablationTab}, ``+6GAP'' means we only embed a GAP module in $\mathcal D^6$, while ``+6GAP\_5AC'' means an AC module is further embedded in $\mathcal D^5$. Other settings can be inferred similarly. The comparison results show that adding GAP in $\mathcal D^6$ can moderately improve the model performance, and progressively embedding AC in latter decoding modules makes further contribution, finally leading to significant performance improvement compared with the baseline model, on all the six datasets and in terms of all the three evaluation metrics.

\begin{figure}[!ht]
  %\graphicspath{{Figures/baselinCmp/}}
  \centering
  \includegraphics[width=1\linewidth]{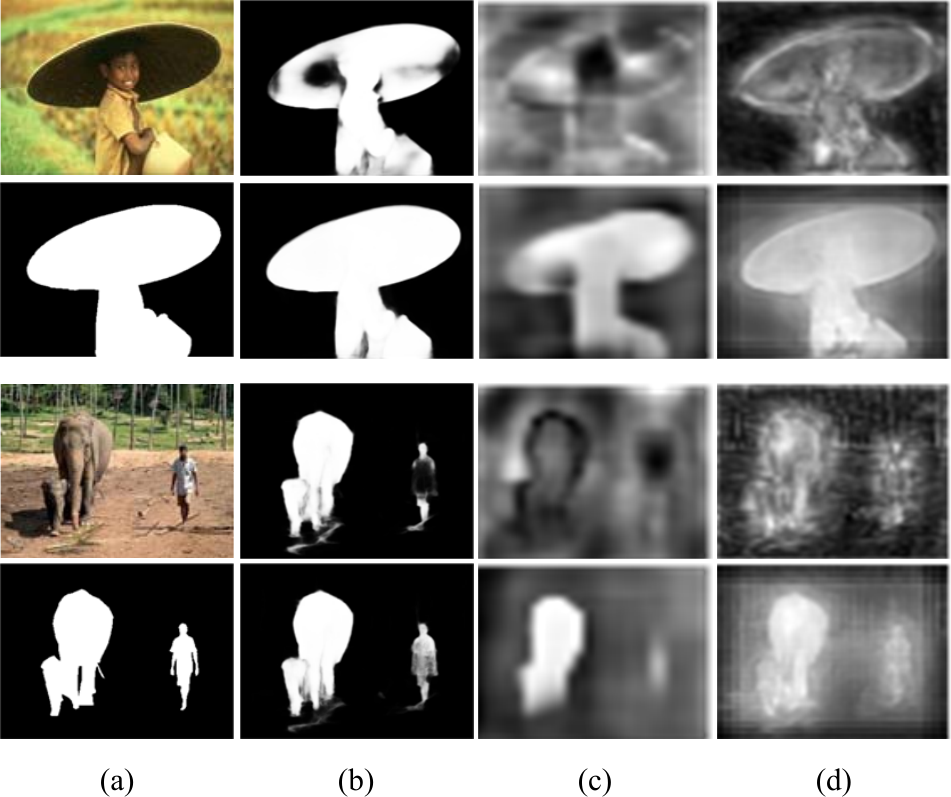}
  \caption{\textbf{Visual comparison of our model against the baseline U-Net}. We show two groups of examples. (a) Two testing images and their ground truth saliency maps. (b) Saliency maps of the baseline U-Net (the top row in each group) and our model (bottom rows). (c) $\bm{F}^6$ (top rows) and $\bm{F}_{att}^6$ (bottom rows). (d) $\bm{F}^2$ (top rows) and $\bm{F}_{att}^2$ (bottom rows).}
  \label{baselinCmp}
  \vspace{-0.3cm}
\end{figure}

\vspace{3mm}
\Paragraph{Different embedding settings.}
We further show comparison results of different embedding settings of our global and local PiCANets, including only adopting local PiCANets (``+65432AC''), and embedding GAP in more decoding modules (``+65GAP\_432AC'' and ``+654GAP\_32AC''). Table~\ref{ablationTab} shows that all these three settings generally performs slightly worse than the ``+6GAP\_5432AC'' setting. We do not consider to use GAP in other more decoding modules since it is time-consuming for large feature maps.

\vspace{3mm}
\Paragraph{AC vs. LAP?}
Both AC and LAP can incorporate attentive local contexts, but which one is better? To this end, we also experiment with using LAP in $\mathcal D^5$ to $\mathcal D^2$ (``+6GAP\_5432LAP''). Compared with the setting ``+6GAP\_5432AC'', Table~\ref{ablationTab} shows that using AC is a slightly better choice for saliency detection.

\vspace{3mm}
\Paragraph{Attention loss?}
The global attention loss $\bm{L}_{GA}^6$ is used to facilitate the learning of the global contrast in GAP and is adopted in all previously discussed network settings. We also evaluate its effectiveness by setting $\gamma^{GA}=0$ to ban this loss term in training, which is denoted as ``+6GAP\_5432AC\_w/o\_$\bm{L}_{GA}^6$'' in Table~\ref{ablationTab}. We can see that this model performs slightly worse than the setting ``+6GAP\_5432AC'', which indicates that using the global attention loss is slightly beneficial. We also experiment with other values for the loss factor $\gamma^{GA}$ and find that the saliency detection performance is not sensitive to the specific value of this factor. Using 0.1, 0.2, 0.3, and 0.4 achieves very similar results.

\begin{figure*}[!t]
  %\graphicspath{{Figures/attention/}}
  \centering
  %\begin{overpic}[width=1\linewidth,grid,tics=2]{attention.jpg}
  \begin{overpic}[width=1\linewidth]{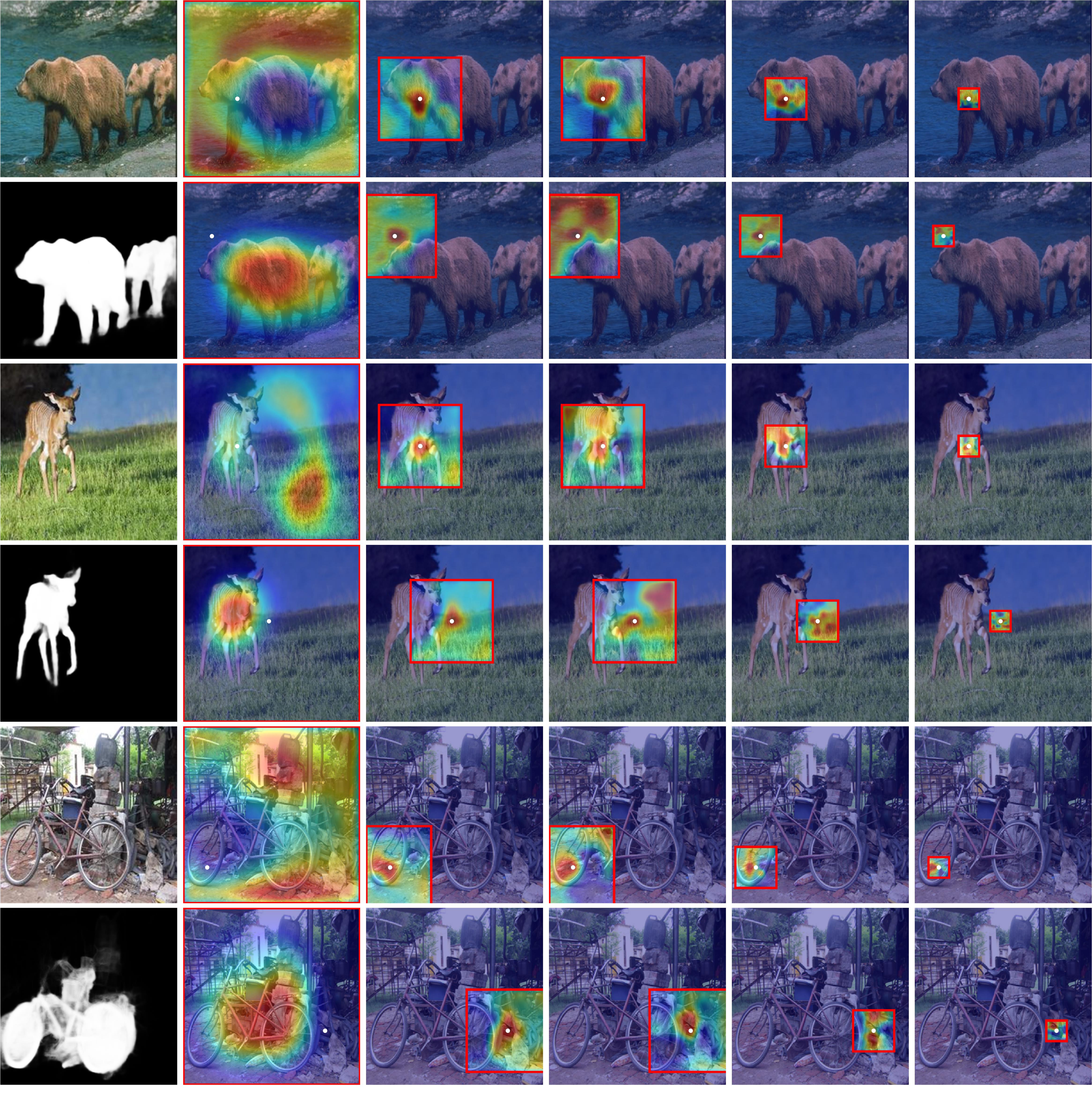}
  \put(1,0){\small Image/Saliency Map}
  \put(21,0){\small att($\mathcal D^6$)}
  \put(38,0){\small att($\mathcal D^5$)}
  \put(54,0){\small att($\mathcal D^4$)}
  \put(70,0){\small att($\mathcal D^3$)}
  \put(86,0){\small att($\mathcal D^2$)}
  \end{overpic}
  \caption{\textbf{Illustration of the generated attention maps of the proposed PiCANets}. The first column shows three images and their predicted saliency maps of our model while the last five columns show the attention maps in five attentive decoding modules, respectively. For each image, we give two example pixels (denoted as white dots), where the first row shows a foreground pixel and the bottom row shows a background pixel. The referred context regions are marked by red rectangles.}
  \label{attention}
  \vspace{-0.4cm}
\end{figure*}
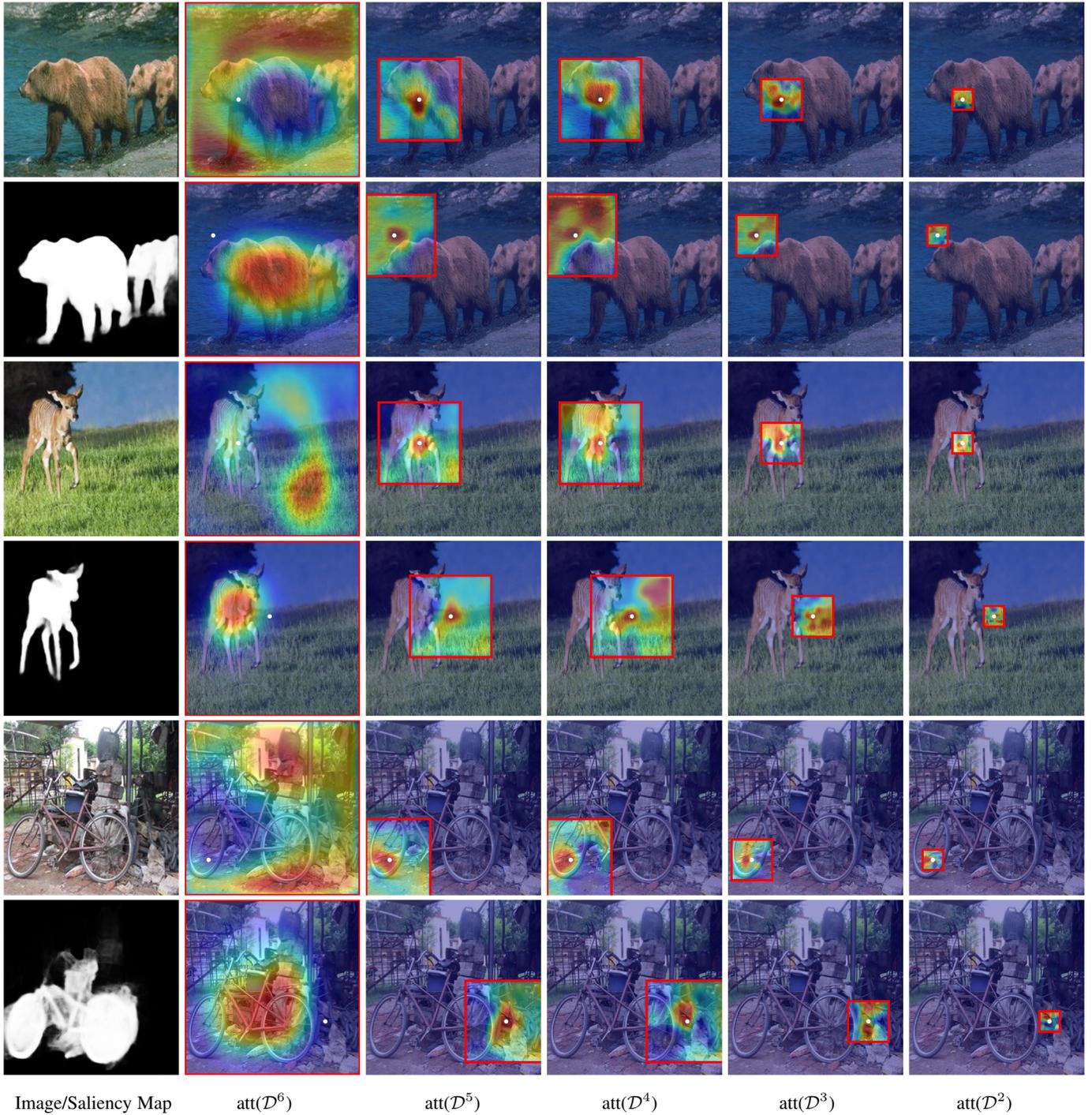

We also experiment with explicitly supervising the training of local attention in AC. Specifically, for each pixel, we use local regions that have the same saliency label with itself as the ground truth attention map and adopt the same KL divergence loss. However, we find that this scheme slightly degrades the model performance. We suppose that this is because regions with the same saliency label do not exactly have similar appearance, especially in cluttered scenes. Thus, the learning of local attention may suffer from the noisy supervision information.

\begin{figure*}[!ht]
  %\graphicspath{{Figures/PR_curves/}}
  \centering
  \includegraphics[width=1\linewidth]{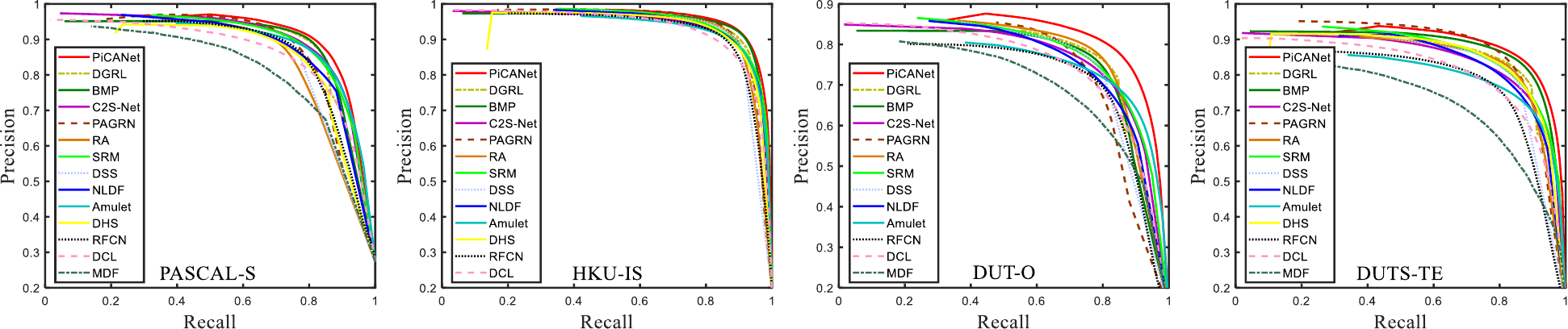}
  \caption{\textbf{Comparison on four large datasets in terms of the PR curve.}}
  \label{prcurve}
  \vspace{-0.1cm}
\end{figure*}

\begin{table*} [!ht]
\begin{center}
\caption{\textbf{Quantitative evaluation of state-of-the-art salient object detection models.} \red{Red} and \blu{blue} indicate the best and the second best performance, respectively.}
\vspace{-0.2cm}
\label{sotaTab}
\footnotesize
\begin{tabular}{@{}R{2cm}|C{0.45cm}C{0.45cm}C{0.45cm}|C{0.45cm}C{0.45cm}C{0.45cm}|C{0.45cm}C{0.45cm}C{0.45cm}|C{0.45cm}C{0.45cm}C{0.45cm}|C{0.45cm}C{0.45cm}C{0.45cm}|C{0.45cm}C{0.45cm}C{0.45cm}}
\toprule
Dataset & \multicolumn{3}{c|}{SOD \cite{yang2013gbmr}} & \multicolumn{3}{c|}{ECSSD \cite{yang2013gbmr}} & \multicolumn{3}{c|}{PASCAL-S \cite{li2014secrets}} & \multicolumn{3}{c|}{HKU-IS \cite{li2015mdf}} & \multicolumn{3}{c|}{DUT-O \cite{yang2013gbmr}} & \multicolumn{3}{c}{DUTS-TE \cite{wang2017duts}} \\ \cmidrule{1-19}
Metric
&$F_{\beta}$&  $S_m$    &   MAE     &$F_{\beta}$&  $S_m$    &   MAE     &$F_{\beta}$&  $S_m$    &   MAE     &$F_{\beta}$&  $S_m$    &   MAE     &$F_{\beta}$&  $S_m$    &   MAE     &$F_{\beta}$&  $S_m$    &   MAE     \\ \midrule
MDF \cite{li2015mdf}
&0.760      &0.633      &0.192      &0.832      &0.776      &0.105      &0.781      &0.672      &0.165      &-          &-          &-          &0.694      &0.721      &0.092      &0.711      &0.727      &0.114
\\
DCL \cite{li2016dcl}
&0.825      &0.745      &0.198      &0.901      &0.868      &0.075      &0.823      &0.783      &0.189      &0.885      &0.861      &0.137      &0.739      &0.764      &0.157      &0.782      &0.795      &0.150
\\
RFCN \cite{wang2016rfcn}
&0.807      &0.717      &0.166      &0.898      &0.860      &0.095      &0.850      &0.793      &0.132      &0.898      &0.859      &0.080      &0.738      &0.774      &0.095      &0.783      &0.791      &0.090
\\
DHS \cite{liu2016dhsnet}
&0.827      &0.747      &0.133      &0.907      &0.884      &0.059      &0.841      &0.788      &0.111      &0.902      &0.881      &0.054      &-          &-          &-          &0.829      &0.836      &0.065
\\
Amulet \cite{Zhang2017amulet}
&0.808      &0.755      &0.145      &0.915      &0.894      &0.059      &0.857      &0.821      &0.103      &0.896      &0.883      &0.052      &0.743      &0.781      &0.098      &0.778      &0.803      &0.085
\\
NLDF \cite{luo2017nldf}
&0.842      &0.753      &0.130      &0.905      &0.875      &0.063      &0.845      &0.790      &0.112      &0.902      &0.879      &0.048      &0.753      &0.770      &0.080      &0.812      &0.815      &0.066
\\
DSS \cite{hou2017dss}
&0.846      &0.749      &0.126      &0.916      &0.882      &0.053      &0.846      &0.777      &0.112      &0.911      &0.881      &0.040      &0.771      &0.788      &0.066      &0.825      &0.822      &0.057
\\
SRM \cite{Wang2017srm}
&0.845      &0.739      &0.132      &0.917      &0.895      &0.054      &0.862      &0.816      &0.098      &0.906      &0.887      &0.046      &0.769      &0.798      &0.069      &0.827      &0.835      &0.059
\\
RA \cite{chen2018ra}
&0.852      &0.761      &0.129      &0.921      &0.893      &0.056      &0.842      &0.772      &0.122      &0.913      &0.887      &0.045      &\blu{0.786}&\blu{0.814}&\red{0.062}&0.831      &0.838      &0.060
\\
PAGRN \cite{zhang2018pagrn}
&-          &-          &-          &0.927      &0.889      &0.061      &0.861      &0.792      &0.111      &0.918      &0.887      &0.048      &0.771      &0.775      &0.071      &\blu{0.855}     &0.837      &0.056
\\
C2S-Net \cite{li2018c2snet}
&0.824      &0.758      &0.128      &0.911      &0.896      &0.053      &0.864      &0.827      &0.092      &0.899      &0.889      &0.046      &0.759      &0.799      &0.072      &0.811      &0.831      &0.062
\\
BMP \cite{zhang2018bmp}
&\blu{0.856}&\blu{0.784}&0.112      &\blu{0.928}&\blu{0.911}&0.045      &\blu{0.877}&\blu{0.831}&\blu{0.086}&\blu{0.921}&\blu{0.907}&\blu{0.039}&0.774      &0.809      &0.064      &0.851      &\blu{0.861}&\red{0.049}
\\
DGRL \cite{wang2018dgrl}
&0.849      &0.770      &\blu{0.110}&0.925      &0.906      &\red{0.043}&0.874      &0.826      &\red{0.085}&0.913      &0.897      &\red{0.037}&0.779      &0.810      &\blu{0.063}&0.834      &0.845      &\blu{0.051}
\\ \midrule
PiCANet (ours)
&\red{0.858}&\red{0.786}&\red{0.107}&\red{0.935}&\red{0.917}&\blu{0.044}&\red{0.883}&\red{0.838}&\red{0.085}&\red{0.924}&\red{0.908}&\blu{0.039}&\red{0.808}&\red{0.835}&0.065      &\red{0.859}&\red{0.867}&\blu{0.051}
\\
\bottomrule
\end{tabular}
\vspace{-0.3cm}
\end{center}{}
\end{table*}

\vspace{3mm}
\Paragraph{Comparison with vanilla pooling and Conv layers.}
Since PiCANets introduce attention weights into pooling and Conv operations to selectively incorporate global and local contexts, we also compare them with vanilla pooling and Conv layers which holistically integrate these contexts for a fair comparison. Specifically, we directly employ the ReNet model \cite{visin2015renet} in $\mathcal D^6$ to capture the global context and use same-sized large Conv kernels (\ie $7\times 7$ kernels with $dilation=2$) in $\mathcal D^5$ to $\mathcal D^2$ to capture the large local contexts, which is denoted as ``+6ReNet\_5432LC'' in Table~\ref{ablationTab}. We also adopt max-pooling (MaxP) and average-pooling (AveP) to incorporate the same-sized contexts, which are denoted as ``+6G\_5432L\_AveP'' and ``+6G\_5432L\_MaxP'', respectively. In $\mathcal D^6$ we first use global pooling and then upsample the pooled feature vector to the same size with $\bm{F}^6$ while in other decoding modules we employ the same-sized local pooling kernels. Compared with the models ``+6GAP\_5432AC'' and ``+6GAP\_5432LAP'', we can see that although using these vanilla schemes to incorporate global and large local contexts can bring moderate performance gains, employing the proposed PiCANets to select informative contexts can achieve better performance.

\vspace{3mm}
\Paragraph{Visual analyses.}
We also show some visual results to demonstrate the effectiveness of the proposed PiCANets. In Figure \ref{baselinCmp}(a) we show two images and their ground truth saliency maps while (b) shows the predicted saliency maps of the baseline U-Net (the top row in each group) and our model (bottom rows). We can see that our saliency model can locate the salient objects more accurately and highlight their whole bodies more uniformly with the help of PiCANets. In Figure \ref{baselinCmp}(c), we show comparison of the Conv feature maps $\bm{F}^6$ (top rows) against the attentive contextual feature maps $\bm{F}_{att}^6$ (bottom rows) in $\mathcal D^6$. While (d) shows $\bm{F}^2$ (top rows) and $\bm{F}_{att}^2$ (bottom rows) in $\mathcal D^2$. We can see that in $\mathcal D^6$ the global PiCANet helps to better discriminate foreground objects from backgrounds, while the local PiCANet in $\mathcal D^2$ enhances the feature maps to be more smooth, which helps to uniformly segment the foreground objects.

To further understand why PiCANets can achieve such improvements, we visualize the generated attention maps of background and foreground pixels in three images in Figure~\ref{attention}. We show the generated global attention maps in the second column. The attention maps show that the GAP modules successively learn global contrast to attend to foreground objects for background pixels and attend to background regions for foreground pixels. Thus GAP can help our network to effectively differentiate salient objects from backgrounds. As for the local attention, since we use fixed attention size ($13\times 13$) for different decoding modules, we can incorporate multiscale attention from coarse to fine, with large contexts to small ones, as shown by red rectangles in the last four columns in Figure~\ref{attention}. The attention maps show that the local attention mainly attends to the regions with the similar appearance with the referred pixel, thus enhancing the saliency maps to be uniform and smooth, as shown in the first column.

%-------------------------------------------------------------------------
\subsection{Comparison with State-of-the-Arts}

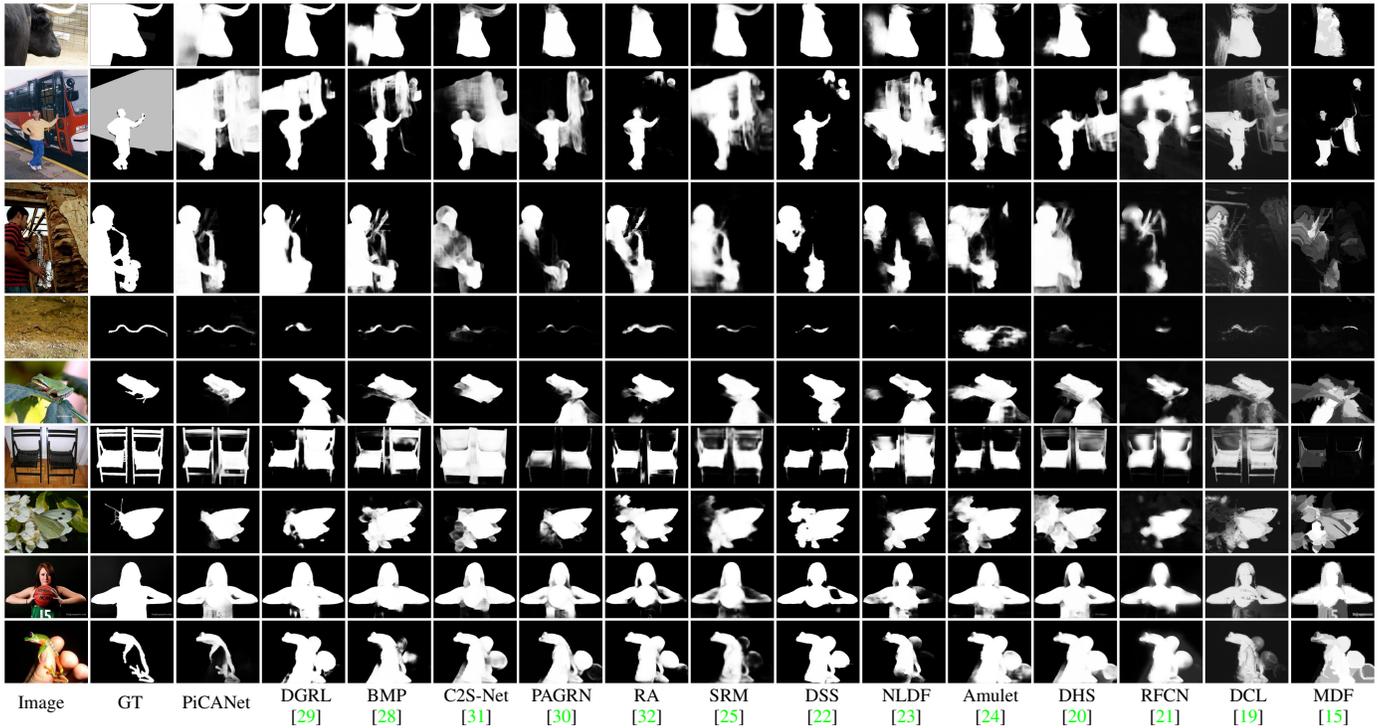
\begin{figure*}[!ht]
  %\graphicspath{{Figures/qualitative/}}
  \centering
  %\includegraphics[width=1\linewidth]{qualitative.jpg}
  %\begin{overpic}[width=1\linewidth,grid,tics=1]{qualitative.jpg}
  \begin{overpic}[width=1\linewidth]{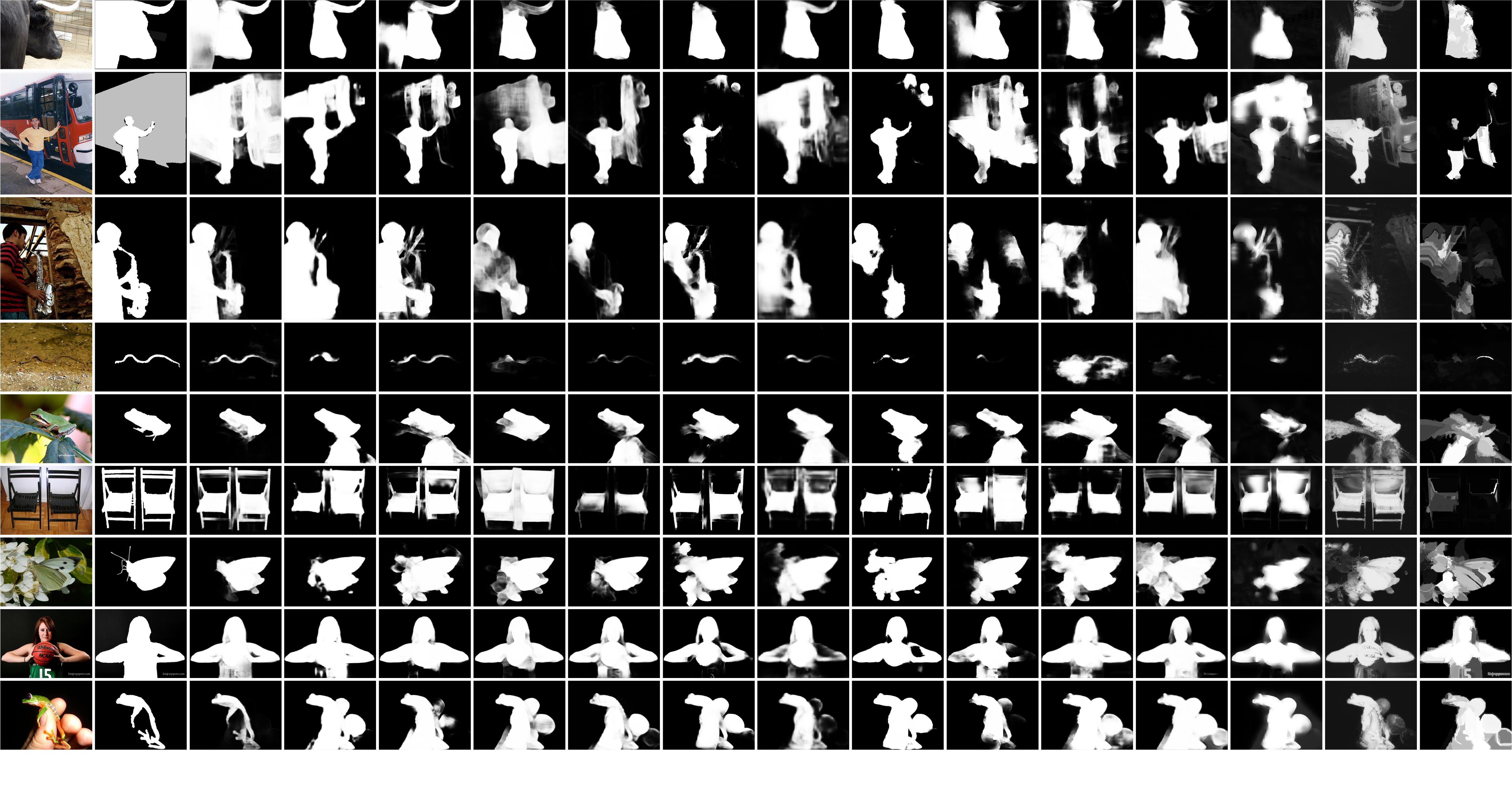}
  \put(1,1){\scriptsize Image}
  \put(8.3,1){\scriptsize GT}
  \put(13,1){\scriptsize PiCANet}
  \put(20.2,0){\scriptsize \shortstack[c] {DGRL\\ \cite{wang2018dgrl}}}
  \put(26.5,0){\scriptsize \shortstack[c] {BMP\\ \cite{zhang2018bmp}}}
  \put(32.1,0){\scriptsize \shortstack[c] {C2S-Net\\ \cite{li2018c2snet}}}
  \put(38.5,0){\scriptsize \shortstack[c] {PAGRN\\ \cite{zhang2018pagrn}}}
  \put(45.8,0){\scriptsize \shortstack[c] {RA\\ \cite{chen2018ra}}}
  \put(51.5,0){\scriptsize \shortstack[c] {SRM\\ \cite{Wang2017srm}}}
  \put(58.5,0){\scriptsize \shortstack[c] {DSS\\ \cite{hou2017dss}}}
  \put(64.1,0){\scriptsize \shortstack[c] {NLDF\\ \cite{luo2017nldf}}}
  \put(70,0){\scriptsize \shortstack[c] {Amulet\\ \cite{Zhang2017amulet}}}
  \put(77,0){\scriptsize \shortstack[c] {DHS\\ \cite{liu2016dhsnet}}}
  \put(83,0){\scriptsize \shortstack[c] {RFCN\\ \cite{wang2016rfcn}}}
  \put(89.5,0){\scriptsize \shortstack[c] {DCL\\ \cite{li2016dcl}}}
  \put(95.6,0){\scriptsize \shortstack[c] {MDF\\ \cite{li2015mdf}}}
  \end{overpic}
  \caption{\textbf{Qualitative comparison with state-of-the-art salient object detection models.} (GT: ground truth)}
  \label{visualcmp}
  \vspace{-0.3cm}
\end{figure*}

Finally we adopt the the setting ``+6GAP\_5432AC'' as our saliency model. To evaluate its effectiveness on saliency detection, we compare it against 13 existing state-of-the-art algorithms, which are DGRL \cite{wang2018dgrl}, BMP \cite{zhang2018bmp}, C2S-Net \cite{li2018c2snet}, PAGRN \cite{zhang2018pagrn}, RA \cite{chen2018ra}, SRM \cite{Wang2017srm}, DSS \cite{hou2017dss}, NLDF \cite{luo2017nldf}, Amulet \cite{Zhang2017amulet}, DHS \cite{liu2016dhsnet}, RFCN \cite{wang2016rfcn}, DCL \cite{li2016dcl}, and MDF \cite{li2015mdf}. All these models are based on deep neural networks and published in recent years. For a fair comparison, we either use the released saliency maps or the codes to generate their saliency maps.\footnote[1]{MDF \cite{li2015mdf} is partly trained on HKU-IS while DHS \cite{liu2016dhsnet} is partly trained on DUT-O. The authors of PAGRN \cite{zhang2018pagrn} did not release the saliency maps on the SOD dataset.}

In Table~\ref{sotaTab}, we show the quantitative comparison results in terms of three metrics. The PR curves on four large datasets are also given in Figure~\ref{prcurve}. We observe that our proposed PiCANet saliency model performs favorably against all other models, especially in terms of the F-measure and the Structure-measure metrics, despite that some other models adopt the conditional random field (CRF) as a post-processing technique or use deeper networks as their backbones. Among other state-of-the-art methods, BMP \cite{zhang2018bmp} and DGRL \cite{wang2018dgrl} belong to the second tier and usually perform better than the rest models.

In Figure~\ref{visualcmp}, we show the qualitative comparison with the selected 13 state-of-the-art saliency models. We observe that our proposed model can handle various challenging scenarios, including images with complex backgrounds and foregrounds (rows 2, 3, 5, and 7), varying object scales, object touching image boundaries (rows 1, 3, and 8), object having similar appearance with the background (rows 4 and 7). Benefiting from the proposed PiCANets, our saliency model can localize the salient objects more accurately and highlight them more uniformly than other models in these complex visual scenes.

%-------------------------------------------------------------------------
\subsection{Application on Other Vision Tasks}

To further validate the effectiveness and the generalization ability of the proposed PiCANets, we also experiment with them on other two dense prediction tasks, \ie semantic segmentation and object detection.

\vspace{3mm}
\Paragraph{Semantic segmentation.}
For semantic segmentation, we first take DeepLab \cite{chen2018deeplab} as the baseline model and embed PiCANets into the ASPP module. Since ASPP uses four $3\times3$ Conv branches with $dilation=\{6,12,18,24\}$, we construct four local PiCANets (\ie AC or LAP modules) with $7\times7$ kernels and $dilation=\{2,4,6,8\}$ to incorporate the same sized receptive fields. Specifically, in each branch, the corresponding local PiCANet is stacked on top of the Pool5 feature map to extract the attentive contextual feature map, which is subsequently concatenated with the Fc6 feature map as the input for the Fc7 layer. Then we train the model by following the training protocols in \cite{chen2018deeplab}. For simplicity, we do not use other strategies proposed in \cite{chen2018deeplab}, \eg, MSC and CRF. For a fair comparison, we also compare PiCANets with vanilla Conv layers with the same large Conv kernels, as denoted by ``+LC''.

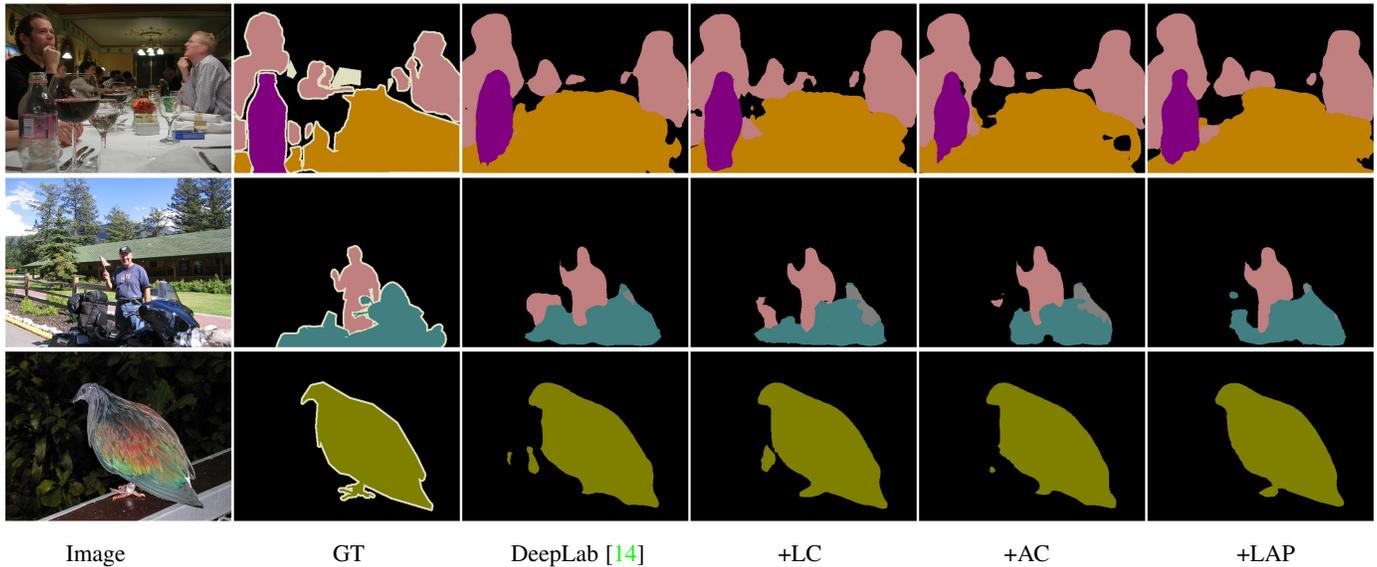
\begin{figure*}[!ht]
  %\graphicspath{{Figures/segmentation/}}
  \centering
  %\begin{overpic}[width=1\linewidth,grid,tics=1]{segmentation.jpg}
  \begin{overpic}[width=1\linewidth]{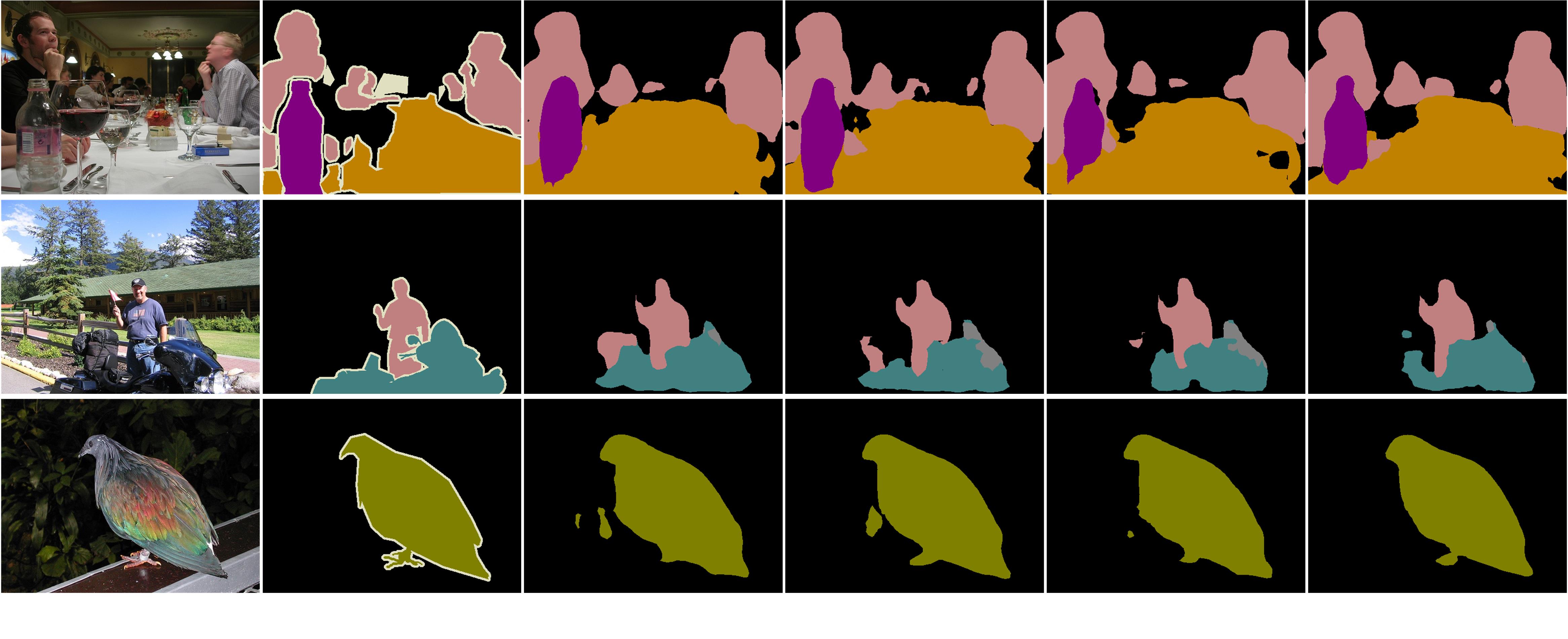}
  \put(4.5,0){\small Image}
  \put(24,0){\small GT}
  \put(37,0){\small DeepLab \cite{chen2018deeplab}}
  \put(56.5,0){\small +LC}
  \put(73,0){\small +AC}
  \put(90,0){\small +LAP}
  \end{overpic}
  \caption{\textbf{Visual comparison of different semantic segmentation model settings.}}
  \label{segcmp}
  %\vspace{-0.1cm}
\end{figure*}

Table~\ref{segTab} shows the model performances on the PASCAL VOC 2012 val set in terms of mean IOU. We observe that as the same as the results on saliency detection, integrating AC and LAP both improve the model performance, while the latter is better for semantic segmentation. Using large Conv kernels here leads to no performance gain, which we believe is probably because their function of holistically incorporating multiscale large receptive fields is heavily overlapped with the ASPP module. This further demonstrates the superiority of the proposed PiCANets. We also give a visual comparison of some segmentation results in Figure~\ref{segcmp}. It shows that using local PiCANets can obtain more accurate and smooth segmentation results by referring to informative neighboring pixels.

\begin{table} [!t]
\begin{center}
\caption{\textbf{Quantitative comparison of different semantic segmentation model settings on the PASCAL VOC 2012 val set in terms of mIOU.} \red{Red} indicates the best performance in each row.}
\vspace{-0.2cm}
\label{segTab}
%\footnotesize
\begin{tabular}{@{}C{1.7cm}|C{1.8cm}|C{1.8cm}|C{1.8cm}}
\toprule
DeepLab \cite{chen2018deeplab}   &+LC             &+AC             &+LAP          \\ \midrule
68.96                            &68.90           &69.33           &\red{70.12}
\\ \toprule
U-Net \cite{ronneberger2015unet} &+6ReNet\_543LC  &+6GAP\_543AC    &+6GAP\_543LAP  \\ \midrule
68.60                            &72.12           &72.78           &\red{73.12}
\\
\bottomrule
\end{tabular}
\vspace{-0.3cm}
\end{center}{}
\end{table}

We also follow the proposed saliency model to adopt the U-Net \cite{ronneberger2015unet} architecture with both global and local PiCANets for semantic segmentation. Generally, the network architecture is similar to the saliency model except that we use $384\times 384$ as the input image size and do not use any dilation in the encoder part. Furthermore, we set the GAP module in $\mathcal D^6$ with the $12\times12$ kernel size and $dilation=1$ and only use the first four decoding modules to save GPU memory. In Table~\ref{segTab}, the comparison results of four model settings show that although adopting ReNet and large Conv kernels can improve the model performance, using PiCANets to select useful context locations can bring more performance gains.

\begin{table} [!t]
\begin{center}
\caption{\textbf{Quantitative comparison of different object detection model settings on the PASCAL VOC 2007 test set in terms of mAP.} ``+478LC\_910ReNet'' means we use vanilla Conv layers with large kernels for the Conv4\_3, FC7, and Conv8\_2 layers and adopt ReNet for the Conv9\_2 and Conv10\_2 layers. Other model settings can be inferred accordingly. \red{Red} indicates the best performance.}
\vspace{-0.2cm}
\label{detectionTab}
%\footnotesize
\begin{tabular}{@{}C{1.1cm}|C{2cm}|C{1.9cm}|C{2cm}}
\toprule
SSD \cite{liu2016ssd}   &+478LC\_910ReNet      &+478AC\_910GAP  &+478LAP\_910GAP \\ \midrule
77.2                       &77.5                  &77.9            &\red{78.0}
\\
\bottomrule
\end{tabular}
\vspace{-0.3cm}
\end{center}{}
\end{table}

\vspace{3mm}
\Paragraph{Object detection.}
For object detection, we leverage the SSD \cite{liu2016ssd} network as the baseline model since it has excellent performance and uses multilevel Conv features which is easy for us to embed global and multi-scale local PiCANets. Specifically, SSD uses the VGG \cite{simonyan2014vgg} 16-layer network as the backbone and conducts bounding box regression and object classification from six Conv feature maps, \ie Conv4\_3, FC7, Conv8\_2, Conv9\_2, Conv10\_2, and Conv11\_2. We deploy local PiCANets with the $7\times7$ kernel size and $dilation=2$ for the first three feature maps and adopt GAP for the latter two according to their gradually reduced spatial sizes. The network structure of Conv11\_2 is kept unchanged since its spatial size is 1. Considering the network architecture and the spatial size of each layer, we make the following network designs:
\begin{compactitem}
\item For Conv4\_3 and FC7, we directly stack an AC module on each of them. Or we can also use the LAP modules, where we concatenate the obtained attentive contextual feature maps with themselves as the inputs for the multibox head.
\item For Conv8\_2, when using LAP, we stack a LAP on top of Conv8\_1 and concatenate the obtained attentive contextual feature with it as the input for the Conv8\_2 layer. When using AC, we directly replace the vanilla Conv layer of Conv8\_2 with an AC module.
\item For Conv10\_2 and Conv11\_2, we deploy a GAP module on each of the Conv10\_1 and Conv11\_1 layers, where the kernel size is set to be equal to the feature map size. Then the obtained attentive contextual features are concatenated with them as the inputs for Conv10\_2 and Conv11\_2.
\end{compactitem}
We also experiment with a model setting to use ReNet and vanilla Conv layers with large kernels to substitute the GAP and AC modules for a fair comparison.

\begin{figure*}[!ht]
  %\graphicspath{{Figures/detection/}}
  \centering
  %\begin{overpic}[width=1\linewidth,grid,tics=1]{detection.jpg}
  \begin{overpic}[width=1\linewidth]{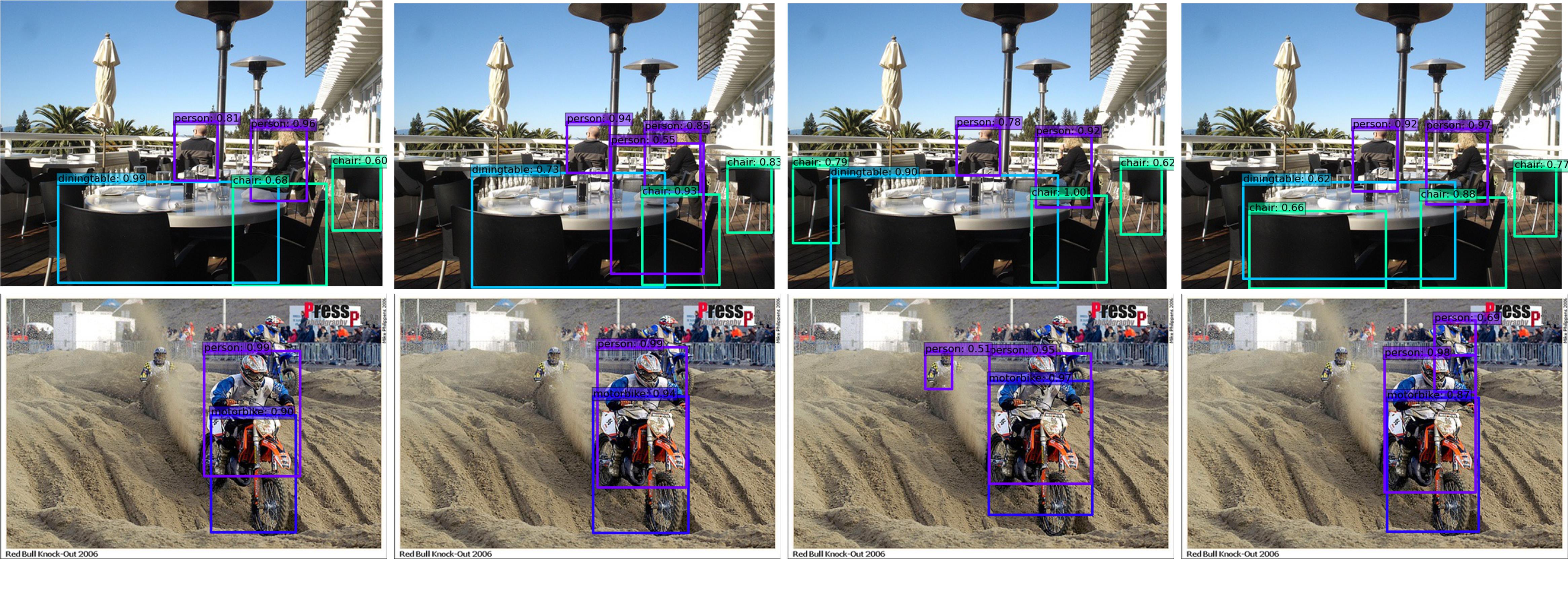}
  \put(8,0){\small SSD \cite{liu2016ssd}}
  \put(29,0){\small +478LC\_910ReNet}
  \put(56,0){\small +478AC\_910GAP}
  \put(81,0){\small +478LAP\_910GAP}
  \end{overpic}
  \caption{\textbf{Visual comparison of different object detection model settings.}}
  \label{detcmp}
  \vspace{-0.3cm}
\end{figure*}
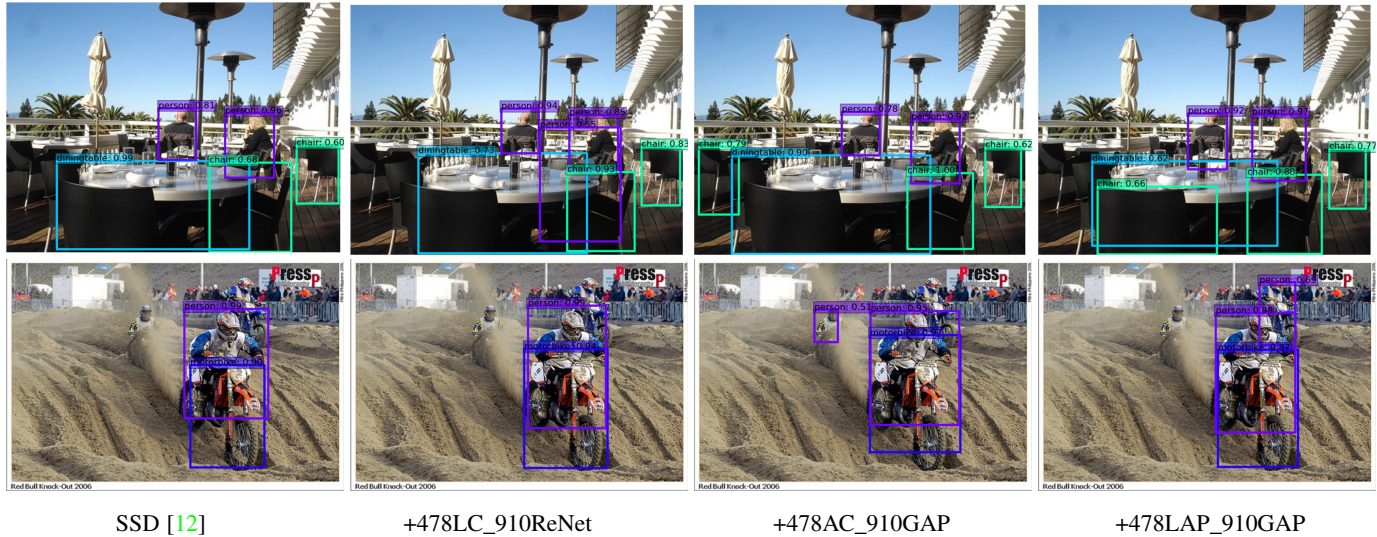

We follow the SSD300 model to use $300\times300$ as the input image size and test on the PASCAL VOC 2007 test set with the mAP metric. The quantitative comparison results are reported in Table~\ref{detectionTab}. It again indicates that using PiCANets with attention can bring more performance gains than the conventional way to holistically incorporate global and local contexts, which is consistent with the previous conclusions. In Figure~\ref{detcmp} we show the detection results of two example images. We can see that PiCANets can either help to generate more accurate bounding boxes, or improve the confidence scores, or detect missing objects.

\section{Conclusion}
In this paper, we propose novel PiCANets to adaptively attend to useful contexts for each pixel. They can learn to generate attention maps over the context regions and then construct attentive contextual features using the attention weights. We formulate the proposed PiCANets into three forms by introducing the pixel-wise contextual attention mechanism into pooling and convolution operations over global or local contexts. All these three modules are fully differentiable and can be embedded into Convnets with end-to-end training. To validate the effectiveness of the proposed PiCANets, we apply them to a U-Net based architecture in a hierarchical fashion to detect salient objects. With the help of the attended contexts, our model achieves the best performance on six benchmark datasets compared with other state-of-the-art methods. We also provide in-depth analyses and show that the global PiCANet helps to learn global contrast while local PiCANets learn smoothness. Furthermore, we also validate PiCANets on semantic segmentation and object detection. The results show that they can bring performance gains on the basis of baseline models, which further demonstrates their effectiveness and generalization ability.

% or


\ifCLASSOPTIONcaptionsoff
  \newpage
\fi



\bibliographystyle{IEEEtran}
\bibliography{PAMI_PiCANet_saliency}
\end{document}